\pdfoutput=1
\documentclass{article}
\usepackage{preprint,times}
\usepackage{amsmath,amsfonts,bm}

\def\eqref#1{equation~\ref{#1}}
\def\1{\bm{1}}

\DeclareMathAlphabet{\mathsfit}{\encodingdefault}{\sfdefault}{m}{sl}
\SetMathAlphabet{\mathsfit}{bold}{\encodingdefault}{\sfdefault}{bx}{n}

\usepackage{hyperref}
\hypersetup{pdftitle={Clean Forcing: Drift-Resistant Autoregressive Video Diffusion with a Frozen Base},pdfauthor={Wenqing Wang, Joonghyuk Shin, Jonathan Tremblay, Chan Hee Song, Yun Fu}}
\usepackage{url}
\usepackage{graphicx}
\usepackage{booktabs}
\usepackage{caption}
\usepackage{multirow}
\usepackage{amsmath,amssymb}
\usepackage{enumitem}
\usepackage{algorithm}
\usepackage{algpseudocode}
\usepackage{tikz}
\usepackage{titletoc}
\usepackage{colortbl}
\usepackage{xcolor}
\usepackage{tabularx}
\usepackage{wrapfig}
\usepackage{needspace}
\usetikzlibrary{calc}

\graphicspath{{figures/}}

\newcommand{\rphi}{r_{\phi}}
\newcommand{\method}{Clean Forcing}

\title{Clean Forcing: Drift-Resistant Autoregressive Video Diffusion with a Frozen Base}

\author{Wenqing Wang$^{1*}$\quad Joonghyuk Shin$^{2}$\quad Jonathan Tremblay$^{3\dagger}$\quad Chan Hee Song$^{3\dagger}$\quad Yun Fu$^{1\dagger}$\\
$^{1}$Northeastern University\quad $^{2}$Seoul National University\quad $^{3}$NVIDIA}

\begin{document}
\maketitle
{\renewcommand{\thefootnote}{}\footnotetext{$^{*}$Work done during an internship at NVIDIA.\qquad $^{\dagger}$Co-last authors.}}

\vspace{-0.32in}
\begin{center}\url{https://clean-forcing.github.io/}\end{center}
\vspace{0.02in}

\begin{figure}[h]
\centering
\includegraphics[width=\linewidth,trim=0 30bp 0 0,clip]{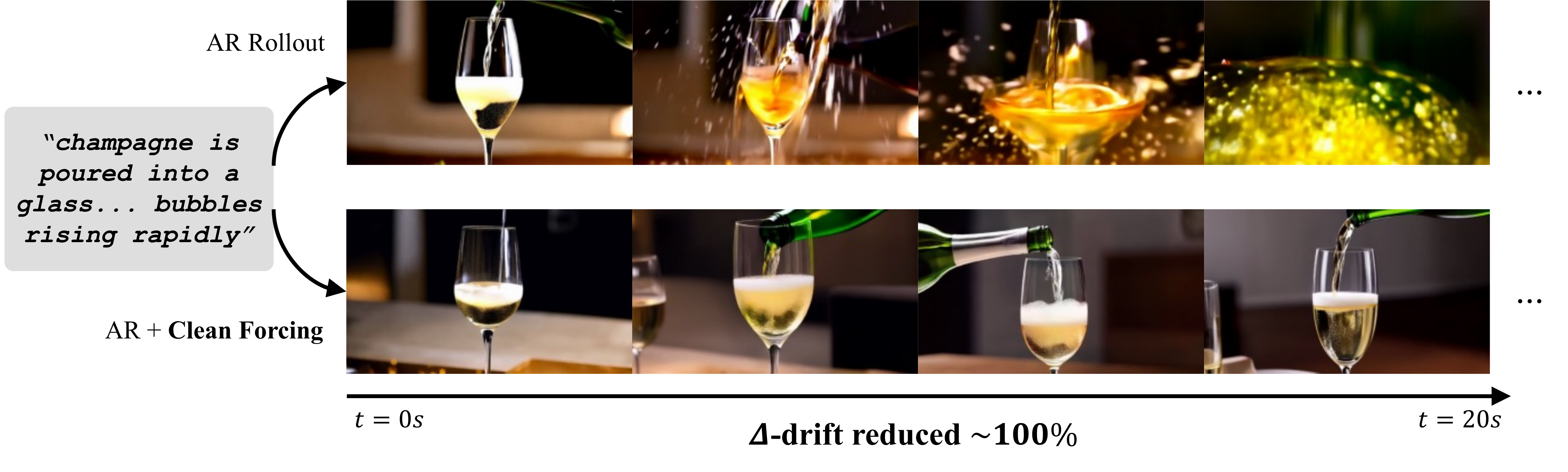}
\caption{\textbf{Clean Forcing reduces drift during AR rollout.}
The first 20\,s of two 50\,s generations with the same prompt and seed are shown. The adapted AR base (top) becomes oversaturated and loses the glass, while the same model with the merged \method{} corrector (bottom) preserves the scene and pouring motion.}
\label{fig:premise}
\end{figure}

\begin{abstract}
    Autoregressive (AR) video diffusion models generate streaming video of arbitrary length by conditioning each new chunk of frames on those already generated. However, they suffer from exposure bias, where a model trained on clean context must continue from its own imperfect outputs, and errors compound until quality collapses within seconds (drift). Existing methods either require cluster-scale base retraining, apply training-free inference corrections with inconsistent results, or train frozen-base correctors for sampling and cache errors. We instead find that drift is largely deterministic and therefore learnable, as holding the noisy state fixed while varying only the history reveals a counterfactual velocity gap that is ${\sim}95\%$ systematic across noise seeds.
    Based on this observation, we introduce \emph{Clean Forcing}, which trains a 5.9M-parameter LoRA corrector on the \emph{frozen} base model to regress this gap, with a closed-loop stage that adds DAgger-style aggregation and a drift-contraction objective to expose the corrector to its own rollouts. Because the clean histories can be the base model's own single-shot generations, training needs \emph{no real videos}, and the merged corrector adds no inference cost. In 50\,s text-to-video generation, Clean Forcing reduces $\Delta$-drift from $+11.3$ to $+1.65$ without real video and to $\mathbf{-0.16}$ with 40 real clips, improving on the best published result of $+3.57$ with over $\mathbf{150\times}$ less data. Extensive experiments demonstrate that Clean Forcing outperforms prior methods in aesthetic quality and data efficiency with reduced drift for long-horizon video generation.
\end{abstract}

\section{Introduction}
Video diffusion models~\citep{wan2025wan,polyak2024moviegen} generate high-quality clips, making them promising for live content creation, simulation, and interactive world modeling. However, most use bidirectional attention~\citep{yin2025causvid} and jointly denoise a clip whose length is fixed before generation begins. Playback therefore cannot begin until the clip has been generated. This inherently limits their applicability to real-time streaming and long-horizon video generation.

Autoregressive (AR) video diffusion models address some of these limitations by generating frames in chunks, with each chunk conditioned on those generated before it~\citep{chen2024diffusionforcing,yin2025causvid,huang2025selfforcing,magi2025,skyreels2025}. This allows rollouts of flexible length and lets playback begin before generation is complete. 
However, models trained to continue from \emph{clean} context must condition on their \emph{own imperfect outputs} at inference. 
Errors are therefore fed back into subsequent chunks and accumulate over time, often causing the video to drift toward oversaturated, textureless frames within seconds (drift; Fig.~\ref{fig:premise}, top). This train--inference mismatch, known as \emph{exposure bias}~\citep{yin2025causvid,huang2025selfforcing,bagger2025}, remains a central challenge for AR video generation.

Existing drift-reduction methods fall into three categories. \emph{Base retraining} incorporates the model's own rollouts into training~\citep{huang2025selfforcing,bagger2025}, which is effective but expensive: the strongest published result uses $136$K clips and $16{\times}$H100 GPUs~\citep{bagger2025}. \emph{Training-free methods} keep the weights fixed and modify inference by noising the context~\citep{chen2024diffusionforcing,skyreels2025}, guiding generation against a cleaner history~\citep{song2025history}, correcting the sampling trajectory~\citep{ttc2026}, or managing the cache~\citep{tethercache2026}. These methods either treat drift as uniform noise or re-anchor generation to a prediction that multi-step models do not produce, and none that we evaluate reduces drift at 50\,s (\S\ref{sec:experiments}). \emph{Frozen-base correctors} learn lightweight corrections, but existing methods target sampling noise~\citep{autorefiner2025} or cache-approximation error~\citep{ertacache2025}, rather than error induced by generated history. Thus, deployed checkpoints still lack an efficient learned correction that directly targets exposure-bias drift.

In this paper, we introduce \textbf{\method{}}, an efficient drift-reduction method for frozen AR video diffusion models. We first characterize history-induced drift by holding a noisy state $z_t$ fixed and comparing the model's denoising velocities under drifted and clean histories. This counterfactual isolates the error induced by the generated history, and our analysis shows that ${\sim}95\%$ of the resulting velocity gap is consistent across noise seeds, indicating that drift is largely systematic and learnable. Based on this insight, we train a rank-16 LoRA corrector $\rphi$ ($5.9$M parameters, $0.4\%$ of the base) to predict this velocity gap while keeping the base model frozen. To expose the corrector to states induced by its own predictions, we introduce a closed-loop stage that regenerates training pairs with the corrector active~\citep{ross2011dagger} and penalizes the downstream drift caused by each committed prediction. The base model's own single-shot generations serve as clean references, so training requires no real videos. On a lightly adapted causal Wan2.1-1.3B~\citep{wan2025wan}, \method{} reduces $\Delta$-drift from $+11.3$ to $-0.16$ over 50\,s on 128 MovieGen prompts~\citep{polyak2024moviegen}. Without real videos, it achieves $+1.65$, outperforming the best published result of $+3.57$~\citep{bagger2025} while using over $150\times$ less data. The corrector trains in one GPU-day and merges into the base model without adding inference overhead.

Our contributions are summarized as follows:
\begin{itemize}[leftmargin=*,itemsep=1pt,topsep=2pt,parsep=0pt]
\item We identify that the velocity error induced by a drifted history is largely systematic across noise seeds, which makes the drift correction learnable.

\item We introduce \method{}, a 5.9M-parameter LoRA corrector trained on a frozen base model to match clean-history predictions at the same noisy state. Training combines a closed-loop stage for DAgger aggregation and drift contraction with a self-teaching procedure that requires no real videos.

\item \method{} eliminates measurable drift using far less data than base retraining and adds no inference overhead. Among the compared methods, it achieves the lowest drift and highest aesthetic score, while ranking second in subject and background consistency, imaging quality, and MUSIQ. It also transfers to an independent base, scales to a $10\times$ larger base, and reduces drift in two other AR world models.
\end{itemize}

\section{Related Work}
\label{sec:related}

\textbf{Video Generation.} Early video generation used GANs~\citep{vondrick2016vgan,tulyakov2018mocogan} or autoregressed over discrete visual tokens~\citep{yan2021videogpt}. GANs offered fast sampling but were difficult to train, while token-based models faced growing generation costs as video sequences lengthened. Diffusion models~\citep{ho2022videodiffusion,blattmann2023svd} improved training stability and visual quality, enabling large-scale systems such as~\citep{brooks2024sora,yang2024cogvideox,kong2024hunyuanvideo,wan2025wan,polyak2024moviegen}. Most of these models jointly denoise a predetermined clip with bidirectional spatiotemporal attention~\citep{peebles2023dit,ho2022imagenvideo,bartal2024lumiere}. Joint denoising promotes temporal coherence, but requires a fixed clip length and delays playback until the full clip is generated, limiting real-time streaming and long-horizon generation.

\textbf{Autoregressive Video Diffusion.} AR diffusion generates video in chunks, enabling streaming and flexible-length rollouts. However, training on clean histories but conditioning on imperfect outputs at inference creates exposure bias, causing errors to compound into drift. To address this mismatch, Diffusion Forcing trains on frames with different noise levels~\citep{chen2024diffusionforcing}, a formulation that DFoT extends to transformers with guidance over the generated history~\citep{song2025history}. Self Forcing takes a different route by training on the model's rollouts~\citep{huang2025selfforcing}. For interactive generation, CausVid distills a bidirectional teacher into a few-step causal student~\citep{yin2025causvid}, an approach further developed by Causal Forcing~\citep{causalforcing2026}. MAGI-1 adds noise to clean context during training to reduce exposure bias~\citep{magi2025}, while SkyReels-V2 uses Diffusion Forcing and noised, overlapping context to improve long-video consistency~\citep{skyreels2025}. These methods mitigate error accumulation without isolating the prediction error caused by the generated history. In contrast, \method{} uses the frozen model's own generations to learn the velocity gap between clean and drifted histories at the same noisy state, reducing drift while preserving streaming, flexible-length rollouts, and inference speed.

\textbf{Mitigating Exposure Bias.} In imitation learning, errors caused by visiting self-generated states are addressed by collecting training examples from the learner's own rollouts~\citep{ross2011dagger}. Video generation methods take a similar approach by retraining the base on its outputs, using distributional targets~\citep{huang2025selfforcing}, reversed rollouts~\citep{bagger2025}, and large-scale rollout training~\citep{rollingforcing2025,selfforcingpp2025,resamplingforcing2025,shin2026motionstream,yang2025longlive}. This improves long-horizon stability, but requires a full model update and $10^4$--$10^5{+}$ clips. Concurrent work post-trains few-step generators on
matched-state targets from real videos~\citep{opsdv2026}. Training-free methods instead noise the context~\citep{chen2024diffusionforcing,skyreels2025}, guide generation with a cleaner history~\citep{song2025history}, adjust the sampling path or activations~\citep{ttc2026,freeloc2026}, or manage the cache~\citep{tethercache2026,deepforcing2025}. They require no training, but do not reliably reduce drift (\S\ref{sec:experiments}). Frozen-base correctors offer a lightweight alternative, but AutoRefiner targets sampling noise~\citep{autorefiner2025}, while ERTACache addresses the separate problem of cache-approximation error~\citep{ertacache2025}. Unlike prior methods, \method{} directly learns the history-induced velocity gap while keeping the base frozen. With at most 340 clips and no real video, it matches or surpasses full retraining in drift reduction and adds no inference cost after merging.

\section{Background}
\label{sec:method}

An autoregressive (AR) video model generates a video as $K$ latent chunks $x=(x^{(1)},\ldots,x^{(K)})$, each containing $f$ frames, conditioned on a text prompt $c$:
\begin{equation}
p_\theta(x\mid c)
=
\prod_{k=1}^{K}p_\theta\!\left(x^{(k)}\mid h^{(k)},c\right),
\qquad
h^{(k)}=x^{(<k)},
\label{eq:ar}
\end{equation}
where $\theta$ denotes the pretrained model parameters and $h^{(k)}$ is the history of committed chunks. Block-causal attention restricts each chunk to itself and its history, whose representations are stored in a KV cache. Each conditional is modeled by flow matching~\citep{lipman2022flow}. Given a clean latent chunk $x_0$, Gaussian noise $\epsilon\sim\mathcal{N}(0,I)$, and noise level $\sigma_t\in[0,1]$ at solver step $t$, the noisy state is $z_t=(1-\sigma_t)x_0+\sigma_t\epsilon$. The network $v_\theta(z_t,h,t)$ predicts the path velocity and is trained with $\mathcal{L}_{\mathrm{FM}}=\mathbb{E}_{x_0,\epsilon,t}\|v_\theta(z_t,h,t)-(\epsilon-x_0)\|_2^2$. At inference, each chunk is initialized from noise and denoised by integrating $\frac{\mathrm{d}z}{\mathrm{d}\sigma}=v_\theta$ from $\sigma=1$ to $0$ over $T$ solver steps, then committed to the history.

\textbf{Exposure bias.}
The model is trained on clean histories but conditioned on its own imperfect outputs at inference, leading to different conditional distributions under the model's generated rollout history~$h_{\mathrm{gen}}$ and the reference history $h_{\mathrm{clean}}$: $p_\theta(x^{(k)}\mid h_{\mathrm{gen}},c) \neq p_\theta(x^{(k)}\mid h_{\mathrm{clean}},c)$~\citep{ranzato2016sequencelevel,schmidt2019exposurebias,ning2023exposurebias,selfforcingpp2025}. Because each generated chunk becomes part of the history, errors compound across $K$ chunks into $O(K^2\varepsilon)$ cumulative error, manifesting as progressive quality drift over longer rollouts~\citep{ross2011dagger}. Our goal is to align these conditional distributions during inference, such that $p_\theta(x^{(k)}\mid h_{\mathrm{gen}},c)\approx p_\theta(x^{(k)}\mid h_{\mathrm{clean}},c)$.

\section{Clean Forcing}
\label{sec:cleanforcing}

\begin{figure}[t!]
\centering
\includegraphics[width=1\linewidth]{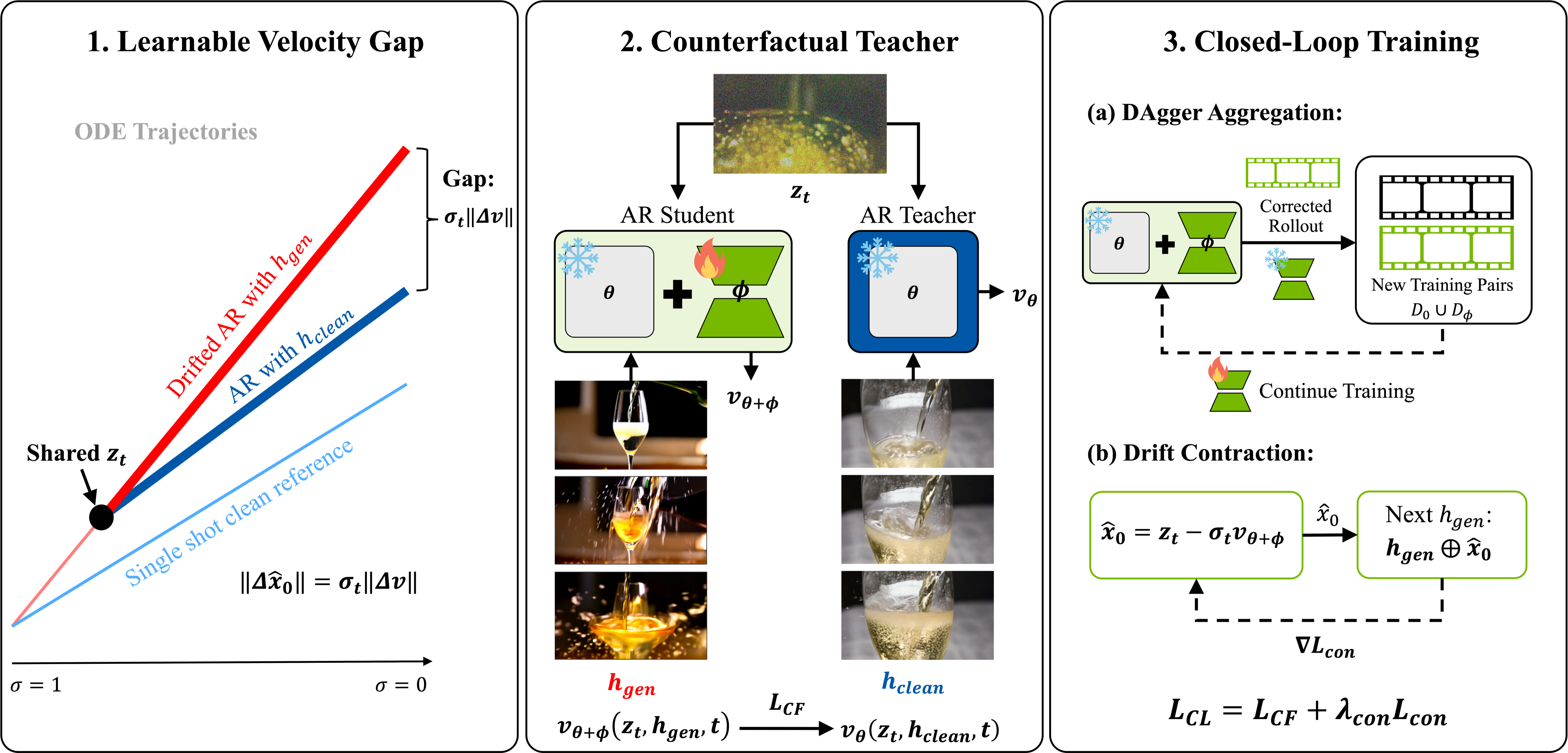}
\caption{
\textbf{Overview.}
\textbf{(1)} At the shared noisy state $z_t$, $\Delta v:=v_\theta(z_t,h_{\mathrm{clean}},t)-v_\theta(z_t,h_{\mathrm{gen}},t)$ is the history-induced velocity gap. The corresponding one-step clean-latent predictions differ by $\sigma_t\|\Delta v\|$, contributing to drift over successive chunks.
\textbf{(2)} The LoRA student conditions on $h_{\mathrm{gen}}$ and matches the frozen teacher's prediction under $h_{\mathrm{clean}}$ ($\mathcal{L}_{\mathrm{CF}}$) ($\phi$ denotes the weights of the LoRA corrector $\rphi$).
\textbf{(3a)} We generate new training pairs $D_\phi$ with the corrector active and continue training on $D_0\cup D_\phi$, where $D_0$ contains pairs from uncorrected rollouts.
\textbf{(3b)} $\mathcal{L}_{\mathrm{con}}$ backpropagates the next chunk’s error through the undetached prediction $\hat{x}_0$.}
\label{fig:pipeline}
\end{figure}

\method{} converts our measurement into a learned corrector in three steps (Fig.~\ref{fig:pipeline}). First, we show that the velocity error induced by a drifted history is largely systematic across noise seeds and can therefore be corrected by a deterministic function (\S\ref{sec:measurement}; panel 1). Second, we train a lightweight LoRA on the frozen base to map the drifted-history velocity to the model's counterfactual clean-history prediction at the same noisy state (\S\ref{sec:v1}; panel 2). Third, we close the train--inference gap through DAgger-style data aggregation and a drift-contraction objective (\S\ref{sec:v2}; panel 3). Finally, we show that clean references need not come from real videos because the same checkpoint can generate its own drift-free supervision in bidirectional mode (\S\ref{sec:sources}).

At inference, the corrected velocity is
\begin{equation}
v_{\mathrm{rect}}
=
v_\theta
+
\alpha\left(v_{\theta+\mathrm{LoRA}}-v_\theta\right),
\label{eq:deploy}
\end{equation}
where $\alpha$ controls the correction strength. Setting $\alpha=1$ recovers the LoRA-merged forward pass and therefore adds no inference overhead. The measured systematic fraction $\alpha^*(t)$ provides a principled correction strength at each solver step. It remains nearly constant at $\approx 0.95$ on our base model, so we use the full merged correction without a time-dependent gating schedule.

\subsection{Drift Is a Systematic Velocity Error}
\label{sec:measurement}

For each state $z_t$ reached by a drifted rollout, we evaluate the frozen model twice while changing only the history:
$\Delta(z_t,t)=v_\theta(z_t,h_{\mathrm{clean}},t)-v_\theta(z_t,h_{\mathrm{gen}},t)$.
Because both predictions use the same state, $\Delta$ isolates the counterfactual effect of history corruption. We measure the fraction of this velocity gap that is reproducible across noise seeds as
\begin{equation}
\alpha^*(t)
=
\frac{\|\mathbb{E}_s[\Delta_s]\|_2^2}
{\|\mathbb{E}_s[\Delta_s]\|_2^2+\operatorname{Var}_s(\Delta_s)},
\label{eq:gate}
\end{equation}
where $s$ indexes noise seeds and
$\operatorname{Var}_s(\Delta_s)=\mathbb{E}_s[\|\Delta_s-\mathbb{E}_s[\Delta_s]\|_2^2]$.
Across all noise levels, the measured systematic fraction $\alpha^*(t)\in[0.89,0.99]$ on three base models (Fig.~\ref{fig:alphagate}). Thus, approximately $95\%$ of the history-induced velocity error is systematic, indicating that a deterministic function can learn to correct it. Because $\alpha^*(t)$ remains close to $1$ throughout denoising, we apply the full correction at every step without a gating schedule, as validated in \S\ref{sec:experiments}. The same checkpoint produces clean clips in bidirectional single-shot mode, which serve as our clean histories (\S\ref{sec:sources}), indicating that drift arises from autoregressive rollout rather than limited model capacity.

\subsection{One-Step Counterfactual-Teacher Correction ($\rphi$)}
\label{sec:v1}

We train the corrector to match the frozen model's clean-history prediction at the same noisy state:
\begin{equation}
\mathcal{L}_{\mathrm{CF}}
=
\mathbb{E}_{k,t,\epsilon}
\left[
\frac{
\left\|
v_{\theta+\mathrm{LoRA}}(z_t,h_{\mathrm{gen}},t)
-
v_\theta(z_t,h_{\mathrm{clean}},t)
\right\|_2^2
}{
\left\|
v_\theta(z_t,h_{\mathrm{clean}},t)
-
v_\theta(z_t,h_{\mathrm{gen}},t)
\right\|_2^2
}
\right].
\label{eq:v1}
\end{equation}
Each training example selects a chunk $x_0$ from a cached drifted rollout and re-noises it as $z_t=(1-\sigma_t)x_0+\sigma_t\epsilon$. The student evaluates $z_t$ under $h_{\mathrm{gen}}$ with the corrector active, while the frozen teacher evaluates the same state under $h_{\mathrm{clean}}$. The denominator normalizes each example by the magnitude of its original drift gap. We report $R^2=1-\mathcal{L}_{\mathrm{CF}}$, the average fraction of each example's drift gap that the corrector closes. The corrector $\rphi$ consists of rank-16 LoRA adapters on the self-attention projections, totaling $5.9$M parameters ($0.4\%$ of the base). We initialize each LoRA output matrix $B$ to zero, so the corrected model initially matches the frozen base exactly.

\textbf{Anchoring.}
Anchoring $z_t$ to a ground-truth trajectory instead can pair the generated history with a state from a different scene, producing a mismatched correction that increases drift in our experiments (\S\ref{sec:experiments}). Under flow matching, the gap between the predicted clean latents scales linearly with the noise level, $\|\Delta\hat{x}_0\|=\sigma_t\|\Delta v\|$ (Fig.~\ref{fig:pipeline}, panel 1).

\subsection{Closed-Loop Training: DAgger and Drift Contraction}
\label{sec:v2}

The one-step objective is trained on states visited by the uncorrected model but deployed on states visited by the corrected model, creating the covariate shift addressed by DAgger~\citep{ross2011dagger}. To close this gap, we augment one-step training with DAgger aggregation and drift contraction.

\textbf{DAgger aggregation.}
We regenerate paired training data with the current corrector active and add it to the existing dataset. Training then samples from this aggregated pool, exposing the corrector to states encountered during its own rollouts.

\textbf{Drift contraction.}
From the corrected velocity, we estimate the clean latent as
\begin{equation}
\hat{x}_0
=
z_t-\sigma_t v_{\theta+\mathrm{LoRA}}(z_t,h_{\mathrm{gen}},t).
\label{eq:commit}
\end{equation}
We append $\hat{x}_0$ to the student history while retaining its gradient, then evaluate the normalized velocity gap on the re-noised next chunk $x_0^{+}$:
\[
\mathcal{L}_{\mathrm{con}}
=
\mathcal{L}_{\mathrm{CF}}
\bigl(x_0^{+};\,h_{\mathrm{gen}}\oplus\hat{x}_0,\,
h_{\mathrm{clean}}\oplus\bar{x}_0\bigr).
\]Here $\oplus$ denotes concatenation, and $\bar{x}_0$ is the clean reference chunk corresponding to $\hat{x}_0$. Backpropagating through $\hat{x}_0$ trains the corrector to reduce the drift its committed prediction causes in the next chunk. The total closed-loop training loss is
\begin{equation}
\mathcal{L}_{\mathrm{CL}}
=
\mathcal{L}_{\mathrm{CF}}
+
\lambda_{\mathrm{con}}\mathcal{L}_{\mathrm{con}},
\label{eq:closed_loop}
\end{equation}
where $\mathcal{L}_{\mathrm{con}}$ denotes the drift-contraction loss and $\lambda_{\mathrm{con}}=0.5$. Ablations in \S\ref{sec:experiments} show that both DAgger aggregation and drift contraction contribute to reducing long-horizon drift.

\subsection{Clean-History Sources}
\label{sec:sources}

The teacher requires clean histories, but these need not come from real videos. We evaluate three settings: (a) \emph{real clips}, for which 40 clips suffice; (b) \emph{prompts only}, where the checkpoint generates single-shot 5\,s clips in bidirectional mode, which avoids autoregressive drift and provides clean self-supervision; and (c) \emph{no clean reference}, where reference-free distribution matching fails to reduce drift (\S\ref{sec:experiments}). Because a bidirectional checkpoint can extend an existing clip but cannot reliably generate from a prompt alone, we causally adapt it using the same synthetic clips and merge the adapter into the base (Appendix~\ref{app:hyperparams}), after which \method{} corrects the remaining drift, with the two providing complementary improvements (\S\ref{sec:experiments}).

\section{Experiments}
\label{sec:experiments}

\textbf{Implementation details.} Our main base is Wan2.1-T2V-1.3B~\citep{wan2025wan}, configured with block-causal attention over 3-latent chunks, a rolling 21-latent KV cache, and a 20-step solver at $832{\times}480$, 16\,fps, CFG $6$, and shift $8$. The \emph{adapted base} has the rank-64 causal adapter from \S\ref{sec:sources} merged into its weights. We also evaluate Wan2.1-T2V-14B, which is $10\times$ larger (Appendix~\ref{app:wan14b}), and an independently trained Causal Forcing checkpoint~\citep{causalforcing2026} (Appendix~\ref{app:cfbase}). The corrector places rank-16 LoRA adapters on the $q$, $k$, $v$, and $o$ self-attention projections in all 30 DiT blocks, yielding $5.9$M trainable parameters ($0.4\%$ of the base). We initialize the LoRA output matrices to zero, use unit scaling, and train with AdamW at learning rates of $5{\times}10^{-4}$ for one-step correction and $2{\times}10^{-4}$ for closed-loop training. The causal adapter uses the same projections and a learning rate of $10^{-4}$. Training uses at most 340 clips (40 real and 300 synthetic); the zero-real-video variant uses only the 300 synthetic clips, versus the $55$K source clips and $81$K generated trajectories of~\citet{bagger2025}. The full pipeline, including generation of teacher clips, takes approximately one day on a single NVIDIA GB300 GPU in bf16 and requires at least 48\,GB of GPU memory (Appendix~\ref{app:hyperparams}).

\textbf{Baselines.} We compare three groups. \emph{(A) Methods implemented on our base}: context noising with $\sigma=0.2$~\citep{chen2024diffusionforcing,skyreels2025}, history guidance with $w=1.2$~\citep{song2025history}, and pathwise test-time correction under both published schedules~\citep{ttc2026}. 
We run these methods on the adapted base using their published hyperparameters. Table~\ref{tab:provenance} documents the implementations and three in-domain controls evaluated on the unadapted base: DMD supervision without clean references~\citep{autorefiner2025,yin2024dmd}, naive fine-tuning, and architecture variants. 
We also evaluate the independently adapted Causal Forcing checkpoint~\citep{causalforcing2026} with and without our correction under the same protocol, since its release does not specify an inference configuration (Table~\ref{tab:cfrobust}). 
\emph{(B) Released checkpoints}: Self Forcing~\citep{huang2025selfforcing} and SkyReels-V2~\citep{skyreels2025} are evaluated on the same 128 prompts. 
\emph{(C) Published results}: BAgger~\citep{bagger2025} is reported on the same calibrated scale but using a different base and prompt set, making direct comparison indicative. 
All groups use the Wan2.1-1.3B architecture (Appendix~\ref{app:fidelity}).

\textbf{Evaluation metrics.} Our main evaluation uses 50\,s generations from 128 LLM-refined MovieGenVideoBench prompts~\citep{polyak2024moviegen} held out from training. All methods use identical prompt and seed pairs (Table~\ref{tab:main}). We report official VBench scores~\citep{huang2024vbench} and $\Delta$-drift, defined as the MUSIQ score~\citep{ke2021musiq} of the first 20\% of frames minus that of the full video~\citep{zhang2025framepack,bagger2025}. To check for metric-specific gains, we also report $\Delta$QualityDrift~\citep{tethercache2026}, quality over time~\citep{yin2025causvid}, color shift, anchored survival~\citep{ttc2026}, and temporal LPIPS (Table~\ref{tab:suite}; Appendix~\ref{sec:guards}). We compare dynamic degree and flicker against ground-truth clips. Error bars show the SEM across prompts. Scoring every frame changes MUSIQ by at most $1.8$ and $\Delta$-drift by at most $0.26$, without changing the $\Delta$-drift ranking. Appendix~\ref{app:metrics} additionally evaluates \emph{in-domain extension} using 5 held-out clips extended by 48 latent frames with 3 paired seeds (Table~\ref{tab:2x2}), and \emph{short-horizon quality} using 200 VBench prompts at 5\,s (Table~\ref{tab:dnh}).

\subsection{Results}
\label{sec:main-results}

\begin{table}[t!]
\centering
    \caption{
    Quantitative comparison on 128 MovieGen prompts at 50\,s, with $\pm$ denoting SEM.
    $\Delta$-drift is first-20\% minus full-video MUSIQ~\citep{bagger2025}.
    Across its variants, \method{} achieves the lowest drift and highest aesthetics.
    It ranks second in subject and background consistency, imaging quality, and MUSIQ.
    Among shared-prompt rows, bold with light red marks the best value and underline with
    light green the second best. Published $\dagger$ values use different prompts, and dynamic degree
    only checks for motion collapse; neither is ranked (Appendix~\ref{sec:guards}).}
\label{tab:main}
\footnotesize
\setlength{\tabcolsep}{2pt}
    \begin{tabular}{@{}lcccccccc@{}}
    \toprule
     & \multicolumn{5}{c}{\textbf{Official VBench $\uparrow$}}
     & \textbf{Motion} & \multicolumn{2}{c}{\textbf{Long-Horizon Quality}} \\
    \cmidrule(lr){2-6}\cmidrule(lr){8-9}
    \textbf{Method} & Subj. & Backg. & Aesth. & Imag. & Smooth
           & Dynamic & MUSIQ$\uparrow$ & $\Delta$-drift$\downarrow$ \\
    \specialrule{\lightrulewidth}{0.4ex}{0pt}
    \rowcolor[gray]{0.93}
    \multicolumn{9}{@{}l@{}}{\strut\textbf{Matched base}} \\
    Adapted base
        & 65.7 & 76.7 & 44.9 & 47.9 & 91.1 & 29.7 & 44.8
        & $+11.27{\scriptstyle\pm0.81}$ \\
    \quad Context noise ($\sigma=0.2$)
        & 67.2 & 77.8 & 45.2 & 51.4 & 85.8 & 32.0 & 49.1
        & $+12.05{\scriptstyle\pm0.76}$ \\
    \quad History guidance ($w=1.2$)
        & 64.6 & 76.1 & 42.4 & 47.7 & 94.5 & 0.8 & 42.3
        & $+20.55{\scriptstyle\pm0.65}$ \\
    \quad Pathwise TTC ($\{500,250\}$)
        & 59.7 & 74.0 & 40.2 & 37.4 & 91.0 & 64.1 & 35.5
        & $+13.96{\scriptstyle\pm0.91}$ \\
    \quad Pathwise TTC ($\{750,500\}$)
        & 61.5 & 75.4 & 42.6 & 39.9 & 91.2 & 71.1 & 38.0
        & $+14.51{\scriptstyle\pm0.80}$ \\
    \textbf{\quad \method{} (zero real videos)}
        & 81.0 & 86.6 & 60.5 & 63.9 & 89.0 & 48.4 & 66.9
        & $+1.65{\scriptstyle\pm0.35}$ \\
    \textbf{\quad \method{} ($+$40 real clips)}
        & 80.9 & 86.4
        & \cellcolor[rgb]{0.90,0.97,0.90}\underline{61.4}
        & \cellcolor[rgb]{0.90,0.97,0.90}\underline{64.1}
        & 89.1 & 49.2
        & \cellcolor[rgb]{0.90,0.97,0.90}\underline{67.0}
        & \cellcolor[rgb]{1.00,0.90,0.90}$\mathbf{-0.16}{\scriptstyle\pm0.12}$ \\
    \rowcolor[gray]{0.93}
    \multicolumn{9}{@{}l@{}}{\strut\textbf{Independent base}} \\
    Causal-Forcing base
        & 64.9 & 75.7 & 45.8 & 56.1 & 90.6 & 33.6 & 52.0
        & $+11.99{\scriptstyle\pm0.67}$ \\
    \textbf{\quad \method{} (zero real videos)}
        & \cellcolor[rgb]{0.90,0.97,0.90}\underline{81.8}
        & \cellcolor[rgb]{0.90,0.97,0.90}\underline{87.9}
        & \cellcolor[rgb]{1.00,0.90,0.90}\textbf{62.9}
        & 62.0 & 91.1 & 70.3 & 62.1
        & $+2.80{\scriptstyle\pm0.42}$ \\
    \rowcolor[gray]{0.93}
    \multicolumn{9}{@{}l@{}}{\strut\textbf{Released models}} \\
    Self Forcing
        & 80.2 & 84.4 & 60.9
        & \cellcolor[rgb]{1.00,0.90,0.90}\textbf{67.7}
        & \cellcolor[rgb]{0.90,0.97,0.90}\underline{94.9}
        & 39.8
        & \cellcolor[rgb]{1.00,0.90,0.90}\textbf{69.3}
        & $+1.08{\scriptstyle\pm0.39}$ \\
    SkyReels-V2-DF
        & \cellcolor[rgb]{1.00,0.90,0.90}\textbf{87.0}
        & \cellcolor[rgb]{1.00,0.90,0.90}\textbf{90.7}
        & 61.0 & 62.8
        & \cellcolor[rgb]{1.00,0.90,0.90}\textbf{96.6}
        & 50.8 & 61.5
        & \cellcolor[rgb]{0.90,0.97,0.90}$\underline{+1.00}{\scriptstyle\pm0.30}$ \\
    \specialrule{\lightrulewidth}{0pt}{0pt}
    \rowcolor[gray]{0.93}
    \multicolumn{9}{@{}l@{}}{\strut\textbf{Published results} ($\dagger$)} \\
    BAgger DF base$^\dagger$
        & 80.7 & 87.7 & 53.1 & 60.0 & 98.1 & 81.3
        & --- & $+7.34$ \\
    BAgger R3$^\dagger$
        & 84.1 & 89.6 & 55.4
        & 63.4 & 98.6 & 76.6
        & --- & $+3.57$ \\
    \bottomrule
\end{tabular}
\end{table}

\textbf{Quantitative results.} Table~\ref{tab:main} validates our claim that a frozen base can match the drift reduction of models fully retrained at cluster scale. On the shared base, \method{} reduces $\Delta$-drift from $+11.27$ to $+1.65$ without real video and to $\mathbf{-0.16}$ with only 40 real clips, indicating no measurable quality decay over 50\,s. Against released, retrained models evaluated on identical prompts, the $+$real variant is lower by more than 3 SEM, while the zero-real variant ties them using over $150\times$ less data. Even the zero-real variant outperforms the best published corrective fine-tuning result on the calibrated scale, with $+1.65$ versus $+3.57$ for R3$^\dagger$. 

\begin{wrapfigure}{rt}{0.48\textwidth}
% \vspace{-14pt}
\centering
\includegraphics[width=\linewidth]{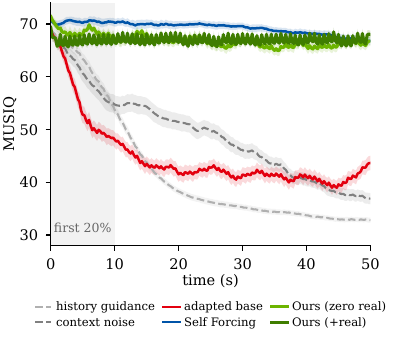}
\caption{{Per-frame quality over 50\,s.}
Mean MUSIQ $\pm$ SEM across 128 prompts. Both \method{} variants remain stable as the base and training-free baselines decline. Ripples mark chunk boundaries.}
\label{fig:money}
% \vspace{-12pt}
\end{wrapfigure}

Drift reduction does not trade off overall quality. Every VBench dimension improves except motion smoothness, which decreases by $2.0$ points as dynamic degree rises from $29.7$ to $49.2$: frame-difference metrics reward stillness, so recovering motion costs smoothness (Appendix~\ref{sec:guards}). On our adapted base, \method{} improves subject consistency and imaging quality by 15--16 points and raises prompt adherence from $58.8$ to $70.3$/$69.9$, comparable to the distilled model's $70.1$ (Table~\ref{tab:semantic}). Its aesthetic score reaches $61.4$ on this base and a table-leading $62.9$ on the Causal-Forcing base. Quality also remains stable when the same videos are evaluated at 10, 30, and 50\,s (Table~\ref{tab:lengths}). SkyReels remains strongest in consistency and smoothness, two dimensions that benefit from large-scale base training.

All training-free interventions we evaluate increase drift relative to the base ($+11.3$): history guidance yields $+20.6$, the two pathwise TTC schedules yield $+14.0$ and $+14.5$, and context noising yields $+12.1$ (Appendix~\ref{sec:guards}). Among the methods in Table~\ref{tab:main}, only \method{} achieves $\Delta$-drift below $+2$ without retraining the base at cluster scale~\citep{huang2025selfforcing,skyreels2025}.

\method{} recipe transfers without model-specific changes or hyperparameter search. On the independently adapted Causal-Forcing base, it reduces drift by $77\%$, from $+11.99$ to $+2.80$, recovers all consistency and aesthetic dimensions to the level of our adapted base, achieves the highest dynamic degree in the table at $70.3$, and restores prompt adherence to $71.0$ (Appendix~\ref{app:cfbase}). At $10\times$ scale, it removes $94\%$ of drift on Wan2.1-14B without changing dynamic degree (Appendix~\ref{app:wan14b}). With a lower correction strength, it also reduces drift in two additional autoregressive world models: HY-WorldPlay~\citep{sun2025worldplay} and the OmniDreams driving model~\citep{basant2026omnidreams} (Appendix~\ref{app:otherhosts}).

\begin{figure}[t!]
\centering
\includegraphics[width=\linewidth]{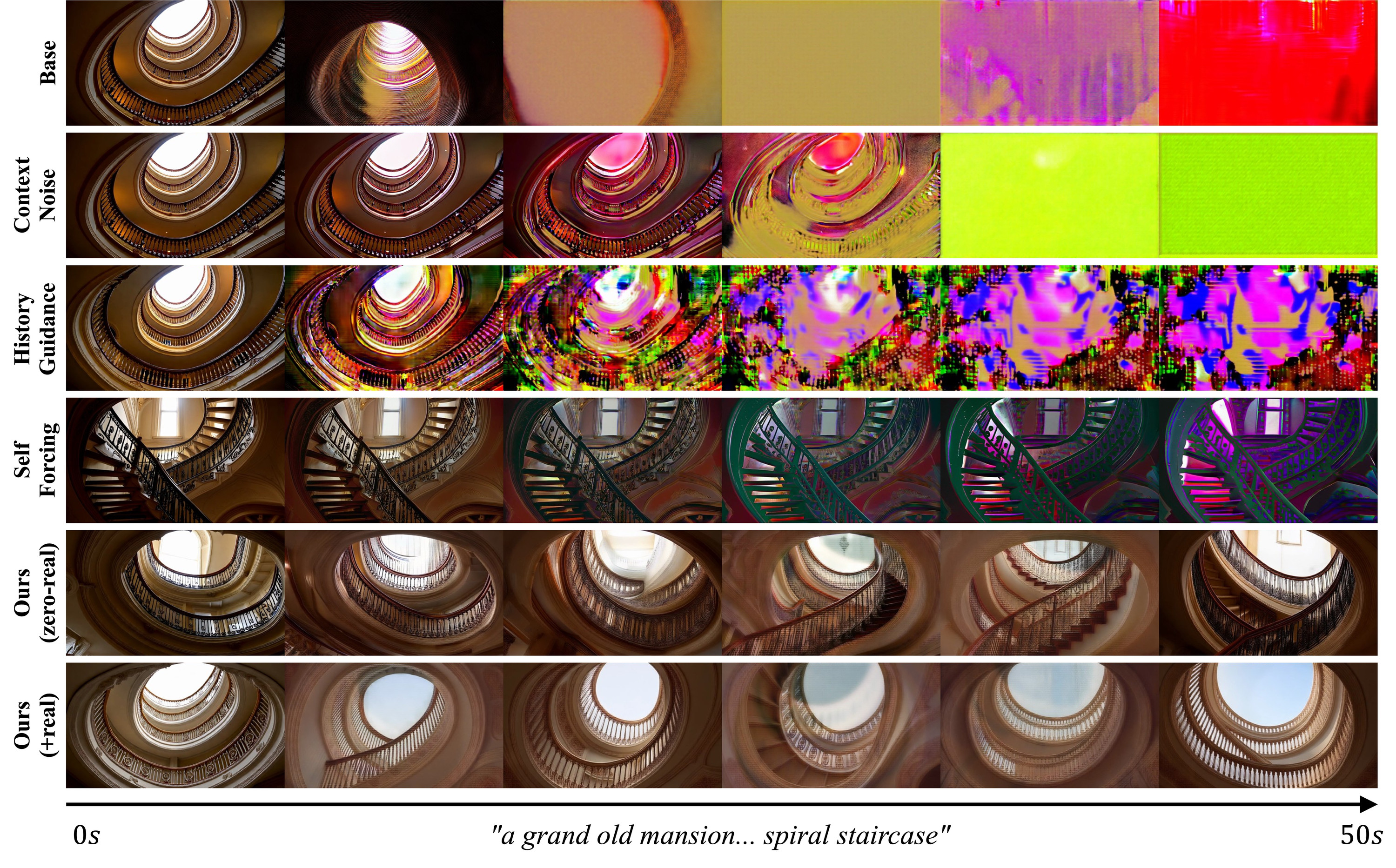}\\[1.5pt]
\includegraphics[width=\linewidth]{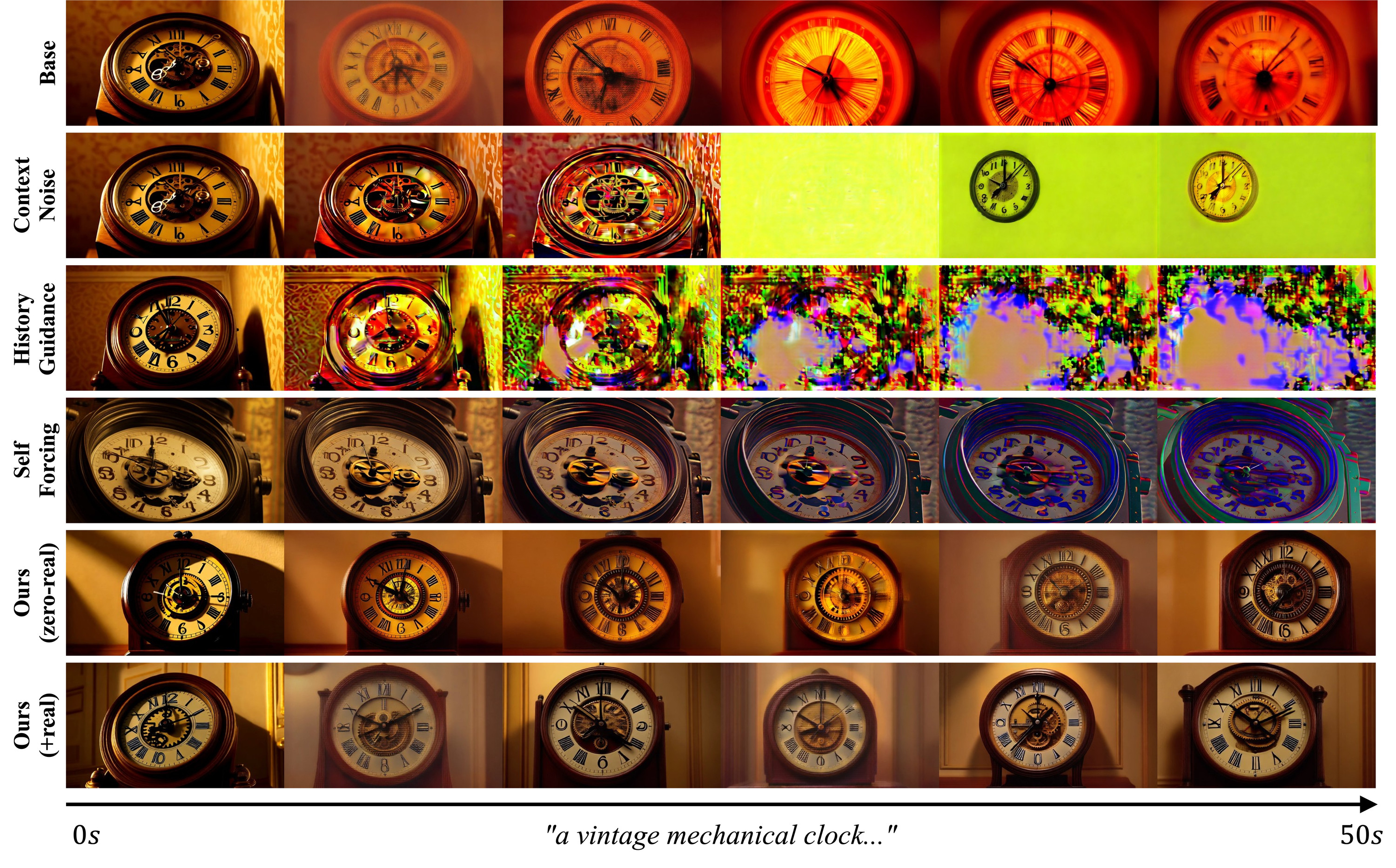}
\caption{{Qualitative comparison over 50\,s.}
Frames are sampled at equal intervals for the spiral staircase (top, p098) and mechanical clock (bottom, p101) prompts. The base and training-free baselines develop saturation or structural artifacts, while Self Forcing shifts color. Both \method{} variants retain the scene throughout the rollout.}
\label{fig:qual}
\end{figure}

\textbf{Qualitative results.} Figure~\ref{fig:money} shows when drift emerges: the base and all training-free heuristics degrade within 15\,s, while both \method{} variants remain stable through 50\,s. Figure~\ref{fig:qual} shows the corresponding frames, in which the base exhibits saturation burnout, the heuristics cause fragmentation or freezing, and Self Forcing develops a late color shift. Appendix~\ref{app:qualitative} presents six additional prompts, Figures~\ref{fig:wan14b_qual} and~\ref{fig:cfstrip} show the same comparison at 14B and on the Causal-Forcing base, and videos for all figure prompts are linked from the project page. Compared with models retrained at cluster scale on the same prompts, \method{} achieves the lowest drift and highest aesthetic score and ranks second in subject and background consistency, imaging quality, and MUSIQ.

\textbf{Efficiency analysis.} At full correction strength, the corrected velocity is exactly the LoRA-merged forward pass, so $\rphi$ folds into the base weights~\citep{hu2022lora} without changing computation, parameter count, peak memory, or NFE ($\Delta\mathrm{NFE}=0$). Wall-clock latency is statistically indistinguishable from the base (Table~\ref{tab:overhead} in Appendix~\ref{sec:deployment}), with a $3.5\%$ increase and Welch $p=0.40$. Merging with a nonconstant schedule $\alpha^*(t)$ is discussed in Appendix~\ref{sec:deployment}. An unfused corrector adds $35\%$ latency, while history guidance doubles NFE and pathwise TTC requires additional denoising passes. Thus, only retraining and \method{} add no inference overhead, and only \method{} is also cheap to train.

\textbf{User study.} Following \citet{yin2025causvid}, 33 raters (29 from Prolific and 4 directly invited) each evaluated 15 randomized pairs of 10\,s late-window excerpts ($t=40$--$50$\,s), with attention checks and at least 5 ratings per pair, yielding 858 judgments. The $+$real variant of \method{} is preferred over the base in $\mathbf{93.0\%}$ of judgments and over Self Forcing in $\mathbf{83.9\%}$, both at $p<10^{-16}$. Its comparison with SkyReels-V2 is not significantly different from an even split ($42.7\%$, $p=0.09$), although SkyReels-V2 is preferred for stability and outputs at 24\,fps versus 16\,fps for all other methods. Table~\ref{tab:userstudy} reports per-question results, and Appendix~\ref{app:userstudy} details the protocol and catch-trial analysis.

\subsection{Ablations}

\begin{table}[t!]
\centering
\caption{{Ablation studies.} We extend held-out clips by 48 latent frames (${\sim}12$\,s), using paired clips and seeds across comparable rows. Bold and underlining indicate the best and second-best reported values. (Drift removed is measured relative to the unadapted base; only this metric was recorded for the DAgger-only checkpoint. Factual targets\textsuperscript{a} collapse motion, making drift and MUSIQ uninformative (Appendix~\ref{sec:guards}). GT-anchored targets\textsuperscript{b} were evaluated on a separate dataset and increased drift, so their values are not directly comparable.)}
\label{tab:ablations}
\small

\begin{tabular}{@{}lccc@{}}
\toprule
Configuration
    & Sat. drift$\downarrow$
    & Drift removed$\uparrow$
    & Late MUSIQ$\uparrow$ \\
\specialrule{\lightrulewidth}{\aboverulesep}{0pt}
\rowcolor[gray]{0.93}[0pt][0pt]
\multicolumn{4}{@{}l@{}}{\strut\textbf{Progressive training}} \\
Unadapted base
    & 0.491 & --- & 29.6 \\
One-step correction
    & \underline{0.166} & 66\% & \underline{36.9} \\
\quad $+$ DAgger aggregation
    & --- & \underline{70\%} & --- \\
\quad $+$ Drift contraction (\textbf{Ours})
    & \textbf{0.131} & \textbf{73\%} & \textbf{60.5} \\
\rowcolor[gray]{0.93}[0pt][0pt]
\multicolumn{4}{@{}l@{}}{\strut\textbf{Alternative targets}} \\
No clean reference (DMD targets)
    & 0.492 & 0\% & 29.6 \\
Factual targets\textsuperscript{a}
    & --- & --- & --- \\
GT-anchored targets\textsuperscript{b}
    & --- & --- & --- \\
\bottomrule
\end{tabular}
\end{table}
Table~\ref{tab:ablations} validates the contribution of each main component of \method{} through progressive additions and individual ablations. We observe that one-step counterfactual correction removes $66\%$ of drift, supporting the clean-history velocity gap as a correction target. Furthermore, the closed-loop components provide further improvements. DAgger raises this to $70\%$, and drift contraction to $73\%$, while late-window MUSIQ rises from $36.9$ to $60.5$. With fixed training data at full scale, closed-loop training reduces $\Delta$ from $+4.77$ to $+1.65$. Additionally, the target ablations explain why the counterfactual matters: reference-free distribution matching removes no drift, factual targets collapse motion, and GT-anchored targets pair different scenes and increase drift. Further ablations cover capacity, architecture, LoRA rank, correction schedules, robustness, scene progression, and checkpoint transfer (Table~\ref{tab:ablations_ext}; Appendix~\ref{app:metrics}; Appendix~\ref{sec:guards}; Appendix~\ref{sec:trade}; Appendix~\ref{sec:deployment}).

\section{Conclusion}
We present \method{}, a 5.9M-parameter LoRA corrector that learns the same-state velocity gap between drifted and clean histories, is refined on its own rollouts, and merges into the frozen base without inference overhead. Because $\sim95\%$ of this gap is systematic, \method{} matches cluster-scale retraining over 50\,s using over $150\times$ less data, while its zero-real-video variant achieves $\Delta$-drift of $+1.65$. Future work could apply the same loss during causal adaptation (Appendix~\ref{app:regularizer}) or combine clean-history supervision with distribution matching to preserve progression. Exposure bias need not be trained away when its effects can be measured and corrected.

\bibliography{refs}
\bibliographystyle{preprint}

\appendix
\clearpage
\section*{Appendices}
The following appendices supplement the main text.

\titlecontents{section}[1.2em]{\vspace{1pt}}{--~Appendix~\thecontentslabel\hspace{0.6em}}{}{}
\titlecontents{subsection}[3.4em]{\vspace{1pt}}{\thecontentslabel\hspace{0.6em}}{}{}
\startcontents[appendix]
\printcontents[appendix]{}{1}{\setcounter{tocdepth}{2}}
\vspace{1em}

\section{Training Details and Hyperparameters}
\label{app:hyperparams}

\begin{algorithm}[h]
\caption{\method{} training}
\small
\begin{algorithmic}[1]
\Require Frozen AR base $v_\theta$; corrector $\rphi$ (LoRA on $v_\theta$, zero-init); weight $\lambda_{\mathrm{con}}$
\Require Clean references $h_{\mathrm{clean}}$ (bidirectional single-shot); drifted rollouts cached from the adapted base
\Loop \Comment{stage 1: one-step}
  \State Sample chunk $x_0$ from a drifted rollout, timestep $t$, $\epsilon \sim \mathcal{N}(0, I)$
  \State Set $z_t \gets (1-\sigma_t)\,x_0 + \sigma_t\epsilon$
  \State Update $\rphi$ on $\mathcal{L}_{\mathrm{CF}}$, the velocity gap at $z_t$ between drifted and clean history \Comment{Eq.~\ref{eq:v1}}
\EndLoop
\State Regenerate rollouts with $\rphi$ active; aggregate into the pool \Comment{DAgger}
\Loop \Comment{stage 2: closed-loop}
  \State Sample from the pool; set $\hat{x}_0 \gets$ corrected prediction \Comment{Eq.~\ref{eq:commit}}
  \State Commit $\hat{x}_0$ to the next chunk's history, with gradient
  \State Update $\rphi$ on $\mathcal{L}_{\mathrm{CF}} + \lambda_{\mathrm{con}}\mathcal{L}_{\mathrm{con}}$ \Comment{Eq.~\ref{eq:closed_loop}}
\EndLoop
\State Merge $\rphi$ into $v_\theta$ at $\alpha{=}1$ \Comment{Eq.~\ref{eq:deploy}; no inference overhead}
\end{algorithmic}
\label{alg:cleanforcing}
\end{algorithm}

All training stages run on one GPU in bf16, with LoRA adapters kept in fp32 and gradient checkpointing enabled. We use AdamW with PyTorch defaults (weight decay $0.01$ and $\beta=(0.9,0.999)$), linear warmup followed by a constant learning rate, gradient clipping at $1.0$, batch size 1, and no EMA. The base uses block-causal attention with 3-latent-frame chunks, a rolling 21-latent KV window, and 20 solver steps. Generation uses CFG $6.0$, shift $8.0$, $832{\times}480$ resolution, and 16\,fps, yielding clips of $21{\times}16{\times}60{\times}104$ latents. Data sampling uses seed 0, while rollout noise is seeded per clip. Table~\ref{tab:hyper} lists the settings for each stage.

\begin{table}[t!]
\centering
\caption{\textbf{Training hyperparameters.}
Training uses AdamW, bf16, batch size 1, linear warmup, and gradient
clipping at $1.0$.}
\label{tab:hyper}
\small
\begin{tabularx}{\linewidth}{@{}l*{4}{>{\raggedright\arraybackslash}X}@{}}
\toprule
 & Causal adaptation & Correction (one-step)
 & Correction (closed-loop) & Teacher clips \\
\midrule
LoRA rank
    & 64 & 16 & 16 (from one-step) & --- \\
Learning rate
    & $10^{-4}$ & $5{\times}10^{-4}$ & $2{\times}10^{-4}$ & --- \\
Warmup steps
    & 100 & 60 & 40 & --- \\
Steps / clips
    & 6,000 (4K deployed) & 800 & 600 & 300 clips \\
Loss / sampling
    & DF velocity MSE
    & Eq.~\ref{eq:v1}
    & Eq.~\ref{eq:closed_loop}
    & 30-step single shot \\
Length
    & 21 latents (7 chunks)
    & 48 (real) / 21 (synth.) latents
    & 48 / 21 latents $+$ 1 contraction chunk
    & 21 latents \\
Data
    & 300 synthetic clips
    & 40 real or 300 synthetic clips
    & Round 0 $+$ one DAgger pool
    & MovieGen prompts \\
\bottomrule
\end{tabularx}
\end{table}

\textbf{Causal adaptation.}
We adapt the base using the Diffusion Forcing objective. For each of the seven chunks, we independently sample $u\sim U(0,1)$, transform it to $\sigma=8u/(1+7u)$, and compute the velocity MSE. We merge the rank-64 LoRA weights into the base after adaptation and reserve 24 of the 300 synthetic clips for validation.

\textbf{One-step corrector.}
At each step, we sample a rollout position $k\in[9,45]$ and a timestep from the 20-step solver schedule. We re-noise the corresponding drifted rollout chunk as $z_t=(1-\sigma_t)x_0+\sigma_t\epsilon$ and optimize the normalized loss in Eq.~\ref{eq:v1}. The clean-history teacher uses the same base network with the corrector LoRA scaled to zero. We initialize the LoRA output matrices $B$ to zero and sample entries of $A$ from $\mathcal{N}(0,1/r)$, so the corrected network initially matches the base. Validation uses $\max(4,N/8)$ held-out clips.

\textbf{Training pairs.} Each pair aligns a drifted rollout with a clean reference by latent position $k\in[9,45]$ ($k\in[9,18]$ for 21-latent synthetic pairs). The adapted base generates the rollout from the reference’s first three latents and caption. For chunk $k$, $h_{\mathrm{gen}}$ and $h_{\mathrm{clean}}$ contain the preceding nine latents from the rollout and reference, respectively, while $\bar{x}_0$ is the aligned reference chunk. References and rollouts span 48 latents for the 40 real clips and 21 latents for synthetic clips. DAgger rollouts reuse the same references, captions, and seeds with the corrector active. In both losses, the denominator uses frozen-base velocities computed without gradients and adds $10^{-8}$ for numerical stability.

\textbf{Closed-loop corrector.}
Starting from the one-step checkpoint, we regenerate paired rollouts with the corrector active at scale 1, reusing the clips and noise seeds from the initial pool. We then sample uniformly from both pools. For drift contraction, we append the corrected estimate $\hat{x}_0=z_t-\sigma_t v_{\theta+\mathrm{LoRA}}$ to the next chunk's history without detaching its gradient and penalize that chunk's normalized velocity gap against the clean teacher. This term has weight $0.5$. We sample positions $k\in[9,42]$ ($k\in[9,15]$ for synthetic pairs) to leave room for the subsequent chunk.

\textbf{Synthetic teacher clips.}
We run the checkpoint in bidirectional single-shot mode with full attention, 30 solver steps, and default guidance. Using MovieGen-bench prompts disjoint from all evaluation prompts and a fixed seed for each clip, we generate 21-latent clips. These clips provide clean histories for the zero-real-video variant and data for causal adaptation.

\section{Baseline Implementation Fidelity}
\label{app:fidelity}
We implement the Group-A baselines on our model because the published methods do not all provide code for this backbone or specify an operation compatible with a deterministic multi-step solver. Table~\ref{tab:provenance} records the source and settings for each implementation, together with its agreement with published qualitative findings. We use the same base and evaluation protocol for the first three methods and evaluate the remaining controls in domain on the unadapted base. These comparisons test the methods in our setting; they do not reproduce every experimental condition of the source papers.

\begin{table}[t!]
\centering
\caption{\textbf{Group-A baseline implementations and outcomes.}
The first three baselines use the 128-prompt, 50\,s protocol
(Table~\ref{tab:main}). The remaining controls use in-domain extensions
on the unadapted base (Tables~\ref{tab:ablations} and~\ref{tab:ablations_ext}).}
\label{tab:provenance}
\small
\begin{tabularx}{\linewidth}{@{}
    >{\raggedright\arraybackslash}p{0.19\linewidth}
    >{\raggedright\arraybackslash}X
    >{\raggedright\arraybackslash}X@{}}
\toprule
Baseline & Implementation & Observed outcome \\
\midrule
Context noise ($\sigma=0.2$)
    & DF-style context noising~\citep{chen2024diffusionforcing,skyreels2025}.
    & Increases drift, consistent with the finding that noising alone
      does not prevent it~\citep{bagger2025}. \\
History guidance ($w=1.2$)
    & DFoT text and history guidance with fractional history masking;
      a simplified variant was also tested~\citep{song2025history}.
    & Neither variant reduces drift, consistent with the replication
      in~\citet{bagger2025}. \\
Pathwise TTC
    & Algorithm~1 of~\citet{ttc2026} with placements
      $\{500,250\}$ and $\{750,500\}$. We re-encode the reference chunk
      at the adjacent RoPE position.
    & Both schedules worsen drift. On our multistep base, the reference
      prediction is blurry when committed as history. \\
\midrule
DMD-LoRA
    & AutoRefiner-style DMD supervision without a clean reference,
      adapted to our deterministic solver~\citep{autorefiner2025,yin2024dmd}.
    & Removes $0\%$ of drift. \\
Naive LoRA fine-tune
    & Flow matching on clean clips, following the $+$LoRA control
      in~\citet{autorefiner2025}.
    & Collapses motion, as reported for that control. \\
Full fine-tune / side network
    & Counterfactual loss with alternative corrector
      parameterizations.
    & Removes $23\%$ / $25\%$ of drift, versus $73\%$ for rank-16 LoRA
      (Table~\ref{tab:ablations_ext}). \\
\bottomrule
\end{tabularx}
\end{table}

The context-noising and history-guidance results agree with the published observation that neither intervention alone resolves long-horizon drift. Our implementations of the remaining controls likewise recover their reported failure modes or isolate the effect of changing the corrector architecture. We released all Group-A implementations and their provenance settings with the code.

The released Group-B checkpoints also differ in training resources. Self Forcing and related DMD approaches train a 1.3B causal generator using signal from a 14B bidirectional teacher alongside a 1.3B fake critic, whereas our corrected models use the frozen 1.3B base as their only teacher. This difference can affect absolute quality and should be considered when comparing checkpoints, although it does not by itself explain the observed performance gap.

\section{Evaluation Checks for Degenerate Outputs}
\label{sec:guards}
A drift metric can improve even when the generated video deteriorates. Several baselines illustrate distinct ways this happens. Naive LoRA fine-tuning obtains subject-consistency and imaging scores of $66.0$ and $58.5$, but its output becomes nearly static: dynamic degree falls to $12.5$ on the finals set and $5.2$ in domain. History guidance reaches $94.5$ motion smoothness at a dynamic degree of only $0.8$. At the opposite extreme, collapsing unadapted rollouts register dynamic degrees of $81$--$100$ because unstable changes count as motion. Pathwise TTC likewise reaches $64$--$71$, roughly $2.2\times$ the base, although its outputs are blurred. That blur also lowers consecutive-frame flicker from $15.9$ for the base to $9.2$ for TTC by making adjacent frames more alike.

We therefore interpret dynamic degree and flicker relative to ground-truth clips, compare quality at matched levels of motion~\citep{ttc2026}, and inspect frame strips for each batch. Reading the metrics together helps distinguish sustained, coherent motion from stillness, blur, and unstable changes.

\section{The Drift--Progression Trade}
\label{sec:trade}

\begin{figure}[t!]
\centering
\includegraphics[width=1\linewidth]{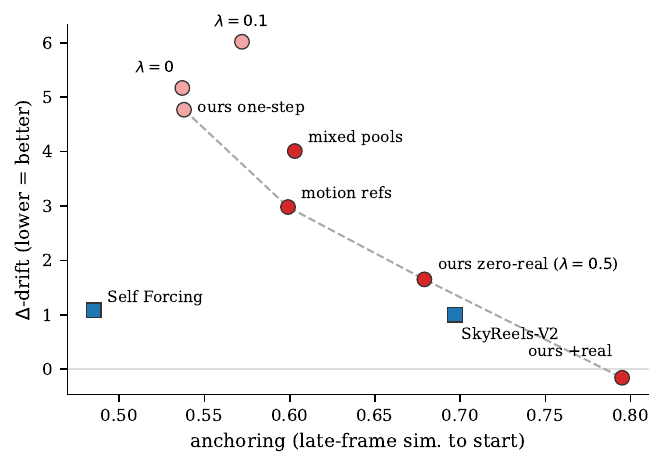}
\caption{\textbf{Drift reduction and scene progression.}
Anchoring is the DINO~\citep{caron2021dino} similarity between the opening and final fifths of a video; higher values indicate less progression. Hollow circles show \method{} variants, filled circles its reported zero-real and $+$real configurations, and squares other systems. Across \method{} settings, lower $\Delta$-drift generally coincides with higher anchoring. Self Forcing, trained on over $10^5$ clips, achieves low drift with less anchoring.}
\label{fig:trade}
\end{figure}
Qualitative inspection revealed a limitation that the main drift score does not capture: corrected scenes can remain close to their opening rather than progress. We measure this tendency as \emph{anchoring}, the DINO-feature similarity between the last fifth of a video and its opening. This measures scene progression and is distinct from the noisy-state anchoring discussed in \S\ref{sec:v1}. We also compare each prompt with Self Forcing to account for differences in the progression implied by the prompt.

Figure~\ref{fig:trade} shows that stronger drift reduction generally coincides with greater anchoring across our variants. One possible mechanism is that drift contraction favors predictions compatible with the clean references, many of which are five-second clips with limited scene changes. Sweeping the contraction weight $\lambda$ moves along this tradeoff, while changing the reference distribution shifts it modestly. SkyReels-V2 lies near the same frontier, with anchoring $0.70$ at $\Delta=+1.00$. Self Forcing attains more progression at low drift after distribution matching on over $10^5$ clips, although the comparison involves a different training recipe and scale. Anchoring alone does not account for our drift reduction: on the less anchored half of each configuration, $\Delta$ remains $+2.2$ for zero real videos and $-0.1$ for $+$real, versus $+11.3$ for the base. Across videos, the correlation between anchoring and drift is $-0.05$.

We also tested whether access to future frames could improve this tradeoff. A bidirectional teacher conditioned on clean past, $z_t$, and clean future reaches anchoring $0.64$ at $\Delta\approx+0.1$, compared with approximately $0.78$ on the measured frontier, but its videos visibly blur. Replacing the clean future with the model's predicted future removes much of the blur and raises anchoring to $0.70$. These results suggest that future information can improve measured progression, but an unpredictable future creates an ambiguous regression target.

A separate limitation is frequent shot changes. Our zero-real videos average about $0.9$ cuts ($0.04$ for $+$real), versus $0.1$--$0.3$ for the retrained models; the uncorrected base averages $2.8$ (Table~\ref{tab:protocol_metrics}). This behavior is partly inherited from the base: extending causal adaptation from 4K to 20K steps roughly halves cuts and flicker and increases MUSIQ by 13 points (Appendix~\ref{app:scaling}). We report anchoring and cut rate alongside drift because neither is captured reliably by the headline quality metrics.

\section{Applicability, Scaling, and Transfer}
\label{sec:deployment}

\begin{wrapfigure}{r}{0.48\textwidth}
\centering
\includegraphics[width=\linewidth]{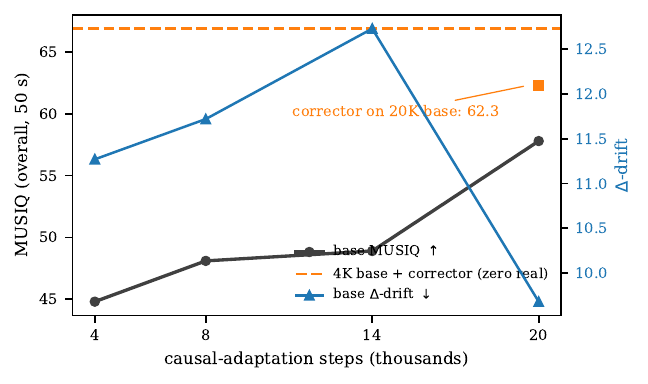}
\caption{\textbf{Longer causal adaptation does not replace correction.}
MUSIQ (black, left axis) improves from 4K to 20K steps, but $\Delta$-drift (blue, right axis) remains substantial. The corrected 4K base reaches 66.9 MUSIQ (dashed line); a corrector retrained on the 20K base reaches 62.3 (square).}
\label{fig:scaling}
\end{wrapfigure}

\textbf{Scaling the base instead.}\label{app:scaling}
Figure~\ref{fig:scaling} evaluates uncorrected bases after 4K--20K causal adaptation steps using the same data and training recipe. Quality continues to improve through 20K, with a substantial gain between 14K and 20K. The corrected 4K base nevertheless reaches $66.9$ MUSIQ, above the uncorrected checkpoints on this curve, with 1,400 corrector updates versus 16,000 additional adaptation steps. Retraining a corrector on the 20K base yields $62.3$ MUSIQ, below the corrected 4K base. Thus, improving the base does not guarantee a better corrected rollout: the corrector must fit the errors of the particular checkpoint to which it is applied.

Increasing the adaptation chunk size from three to seven latents also improves the uncorrected base at the same training budget, giving late MUSIQ $54.2$ and $0.25$ cuts per video. It remains below the corrected three-latent base at approximately $67.0$ late MUSIQ, and the periodic chunk artifact persists at the new cadence. Longer adaptation and larger chunks improve base quality, but neither eliminates the measured drift on its own.

\textbf{Checkpoint transfer.}
When we apply a corrector to a sibling checkpoint, late MUSIQ remains $56.6$, compared with $59.4$ on its native checkpoint, but it retains only about $25\%$ of its drift calibration. Transfer across model types is weaker: a corrector trained on a multi-step velocity model gives $R^2=+0.03$ on a distilled four-step $x_0$ model, compared with $+0.39$ on its source. The training recipe transfers to other bases, but the learned weights are specific to the base whose error they correct.

\textbf{Within-family evidence.}
The same recipe yields a weaker corrected result on a base trained for five times as many adaptation steps, despite that base having less drift before correction. A trajectory-conditioned probe also fails to explain additional gap variance beyond the current state under cross-validation with swept regularization. Together, these results indicate that the correctable signal depends on the particular base and is not recovered simply by adding trajectory context.

\begin{table}[t!]
\centering
\caption{\textbf{Inference overhead.}
Timings use three runs of 21 steady-state chunks on one GPU (bf16,
batch 1), with matched prompt and seed. Chunk latency excludes decoding;
decoded fps includes it. Merging adds no NFE or peak memory, with no
significant latency difference from the base (Welch $p=0.40$).
First-frame latency is unchanged.}
\label{tab:overhead}
\small
\begin{tabular}{@{}lcccccc@{}}
\toprule
 & \multicolumn{2}{c}{Chunk latency}
 & \multicolumn{2}{c}{Compute}
 & \multicolumn{2}{c}{Throughput (fps)} \\
\cmidrule(lr){2-3}\cmidrule(lr){4-5}\cmidrule(lr){6-7}
Method & s/chunk$\downarrow$ & vs.\ base
       & $\Delta$NFE & Peak VRAM (GiB)
       & DiT$\uparrow$ & With decode$\uparrow$ \\
\midrule
Adapted base
    & $2.90\pm0.35$ & ---
    & --- & 29.49 & 4.10 & 3.88 \\
\midrule
\textbf{Merged corrector}
    & $3.00\pm0.89$ & $+3.5\%$
    & 0 & 29.49 & 4.01 & 3.80 \\
Unfused corrector
    & $3.92\pm0.49$ & $+35.3\%$
    & 0 & 29.51 & 3.02 & 2.91 \\
\bottomrule
\end{tabular}
\end{table}

\textbf{Correction schedules.}
Eq.~\ref{eq:deploy} blends model outputs, whereas merging a scaled LoRA update yields $v_{\theta+\alpha\Delta W}$. Because the network is nonlinear in its weights, these expressions generally differ for $0<\alpha<1$; they coincide at $\alpha\in\{0,1\}$. All main results use $\alpha=1$. Timestep-dependent strength $\alpha^*(t)$ requires no additional forward pass when a merged weight set is stored for each distinct strength. The reduced strengths in Appendix~\ref{app:otherhosts} use this implementation.

\textbf{When correction does not help.}
We applied the full recipe to Self Forcing's distilled generator, using an $x_0$-space loss at its four distillation timesteps. The one-step objective was learnable ($R^2=0.48$), but the deployed corrector reduced MUSIQ from $69.3$ to $47.5$ and increased $\Delta$ from $+1.08$ to $+8.3$; DAgger did not reverse this deterioration. Before training, the diagnostic showed a smaller systematic fraction, $\alpha^*(t)\in[0.65,0.80]$, than on the bases where correction helped ($0.89$--$0.99$), and the velocity gap itself was about ten times smaller. A modest fitting error can therefore outweigh the remaining correctable signal when a short solver commits each prediction.

\begin{figure}[t!]
\centering
\includegraphics[width=1\linewidth]{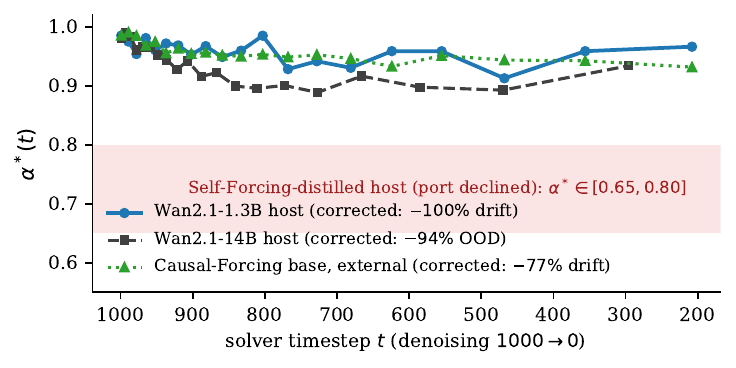}
\caption{\textbf{Systematic velocity gap across bases.}
Per-timestep $\alpha^*(t)$ (Eq.~\ref{eq:gate}) is high on the three bases that benefit from correction (markers). The Self Forcing distilled generator has a lower measured range, $[0.65,0.80]$ (shaded band), and its corrected rollout deteriorates. All measurements precede corrector training.}
\label{fig:alphagate}
\end{figure}

The product of the measured gap magnitude and its systematic fraction provides a practical indicator of potential benefit, rather than a guarantee. It identifies lightly adapted bases with a large, reproducible history-induced error as candidates for correction and suggests limited headroom on a distilled base that already has little drift. Figure~\ref{fig:alphagate} plots the per-timestep fractions for these bases. The observed failure of the distilled port is consistent with the possibility that large-scale distillation has already reduced much of the systematic gap, although these experiments do not isolate that mechanism.

Concurrent results on related distilled models report further gains after generator updates using real-video signal~\citep{opsdv2026}. This is consistent with limited room for a frozen-base corrector after distillation, though those results use a different training method.

\section{Clean Forcing as a Training Regularizer}
\label{app:regularizer}
The counterfactual loss could also be incorporated into causal training. If training already includes on-policy student rollouts, as in Self Forcing-style recipes, the following term would compare the student's prediction under its generated history with a slowly updated teacher's prediction under a clean history:

\begin{equation}
\mathcal{L} \;=\; \mathcal{L}_{\mathrm{FM}} \;+\; \lambda_{\mathrm{CF}}\,
\mathbb{E}\,\frac{\left\lVert v_\theta(z_t, h_{\mathrm{gen}}^{\theta}, t) -
\mathrm{sg}\!\left[v_{\bar\theta}(z_t, h_{\mathrm{clean}}, t)\right]\right\rVert^2}
{\left\lVert v_{\bar\theta}(z_t, h_{\mathrm{clean}}, t) - v_{\bar\theta}(z_t, h_{\mathrm{gen}}^{\theta},
t)\right\rVert^2},
\label{eq:regularizer}
\end{equation}
Here $h_{\mathrm{gen}}^{\theta}$ comes from a short rollout of the current student, $\bar\theta$ denotes an EMA teacher, and $\mathrm{sg}$ stops gradients through the target. The EMA could stabilize the clean-history target as the student changes. This formulation aims to reduce the conditional mismatch during base training; we have not evaluated it experimentally. It may fit naturally into training loops that already compute student rollouts and teacher predictions. Used on its own, however, it requires rollouts during base training and would incur additional cost. Our reported experiments use the corrector trained after the base is fixed; Eq.~\ref{eq:regularizer} is a proposed extension.

\section{Transfer to Other Bases}
\label{app:transfer}

\subsection{Scale: Wan2.1-14B}
\label{app:wan14b}
We apply the zero-real-video recipe to Wan2.1-T2V-14B, a model with roughly ten times the parameters, using the same protocol: 300 synthetic teacher clips, rank-64 causal adaptation, one-step correction, and one round of DAgger followed by drift contraction. The complete pipeline takes about three GPU-days.

\textbf{Diagnostic and adaptation.}
The unadapted 14B base collapses within its first two generated chunks. Its counterfactual velocity gap remains predominantly systematic, with pooled $\alpha^*=0.934$, $\alpha^*=0.961$ at $t\ge900$, and a relative gap of $0.205$. Following the same checkpoint selection rule as at 1.3B, we deploy step 4K from a 6K adaptation run. This checkpoint has a saturation ramp of $0.035$ and late-window MUSIQ of $65.0$.

\textbf{Corrector training.}
The one-step corrector reaches validation $R^2=0.482$. The closed-loop stage peaks at $0.643$ before falling to $0.415$ at its final checkpoint, so we deploy the validation-peak snapshot. This oscillation makes checkpoint selection consequential at 14B.

\begin{figure}[p]
\centering
\includegraphics[width=\linewidth,height=0.40\textheight,keepaspectratio]{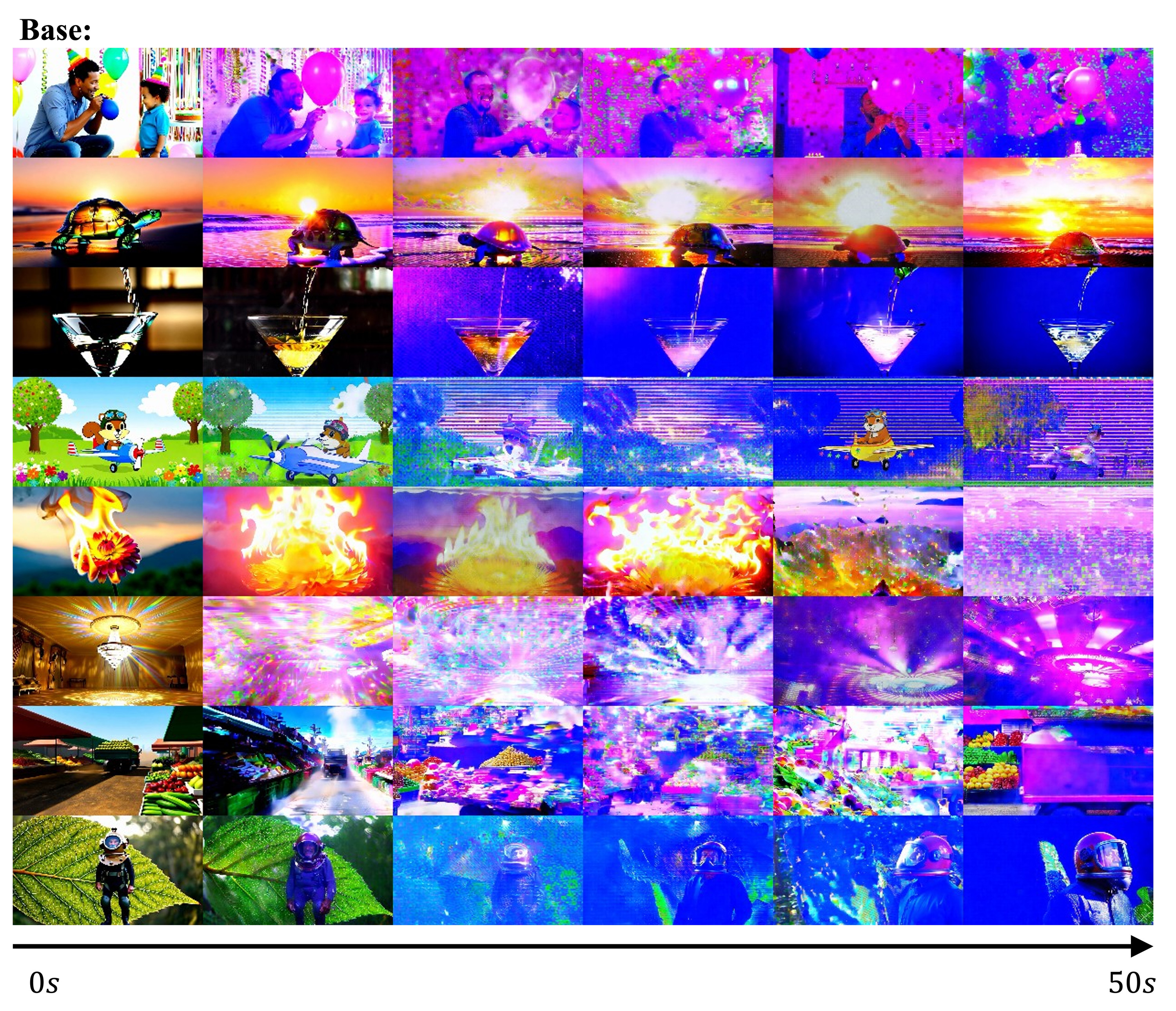}
\par\vspace{2pt}
\includegraphics[width=\linewidth,height=0.40\textheight,keepaspectratio]{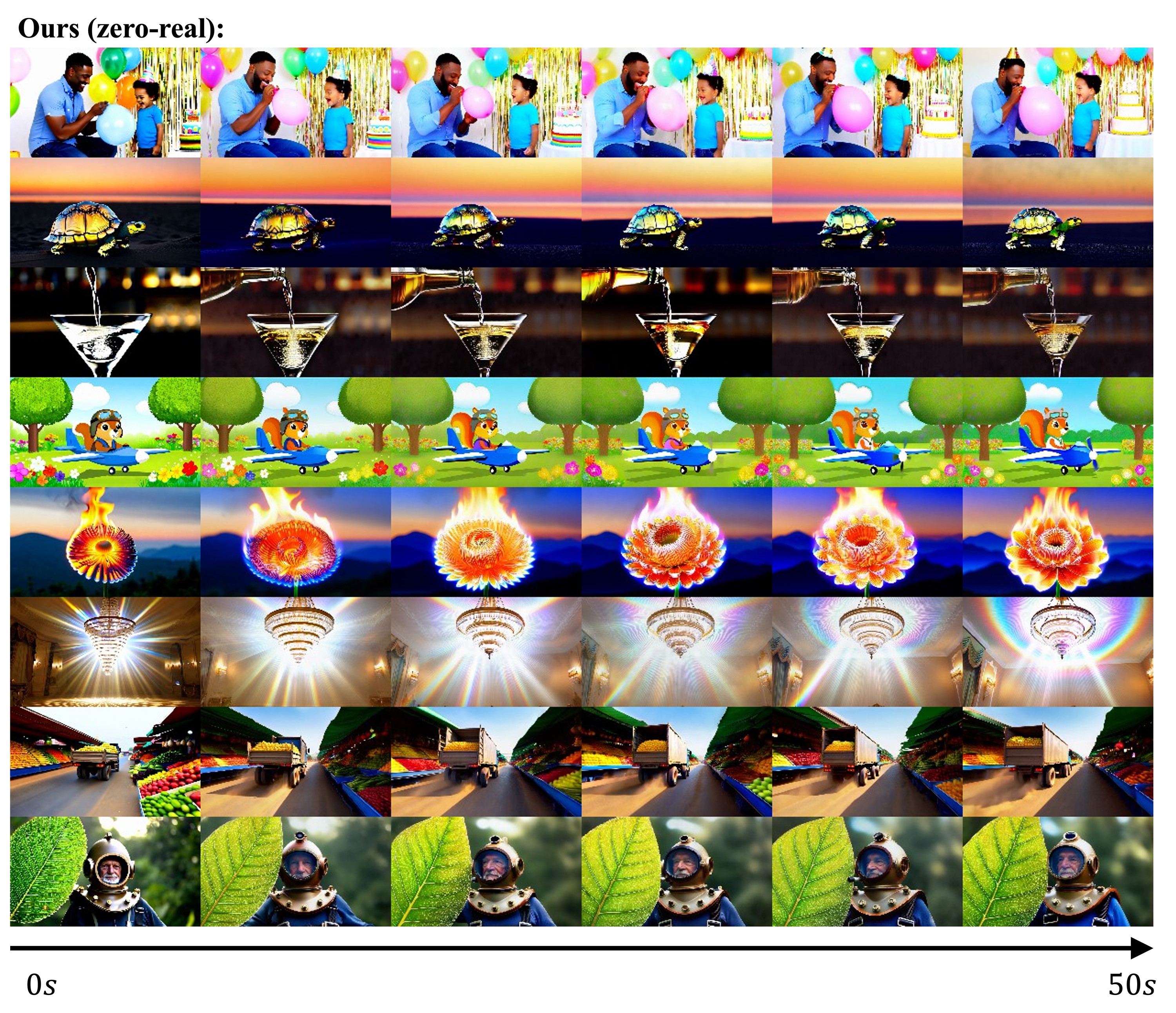}
\caption{\textbf{Qualitative comparison on Wan2.1-14B.}
Six frames sampled over 50\,s for eight held-out prompts, with seeds matched between the adapted base (top) and the same base with the zero-real \method{} corrector merged (bottom). The base progressively saturates and loses scene structure; the corrected model preserves the scenes.}
\label{fig:wan14b_qual}
\end{figure}

\textbf{Results.}
On the paired in-domain extension task (Table~\ref{tab:wan14b}), correction reduces saturation drift by $67$--$70\%$ on the unadapted 14B and $60$--$63\%$ on the adapted base, with gains of up to 43 points in late-window MUSIQ. On 32 held-out MovieGen prompts at 50\,s, the closed-loop corrector reduces $\Delta$-drift by $94\%$, from $+15.65$ to $+0.97$, and raises MUSIQ from $46.4$ to $69.0$ (Table~\ref{tab:wan14b_ood}; Fig.~\ref{fig:wan14b_qual}). Every reported VBench dimension improves while paired dynamic degree remains $37.5$, supporting drift reduction without a loss of measured motion.

\begin{table}[t!]
\centering
\caption{\textbf{Wan2.1-14B in-domain extension.}
Five held-out clips are extended by 48 latent frames under three paired
seeds. Both correctors use the zero-real data. Parentheses give drift
removed relative to the corresponding base; the closed-loop checkpoint
is selected by validation performance.}
\label{tab:wan14b}
\small
\begin{tabular}{@{}lcccc@{}}
\toprule
 & \multicolumn{2}{c}{Unadapted 14B}
 & \multicolumn{2}{c}{Adapted 14B} \\
\cmidrule(lr){2-3}\cmidrule(lr){4-5}
Method & Sat. drift$\downarrow$ & Late MUSIQ$\uparrow$
       & Sat. drift$\downarrow$ & Late MUSIQ$\uparrow$ \\
\midrule
Base
    & 0.329 & 28.4
    & 0.147 & 38.7 \\
\midrule
$+$ One-step corrector
    & 0.109 (67\%) & \textbf{72.1}
    & 0.059 (60\%) & \textbf{71.5} \\
$+$ Closed-loop corrector
    & \textbf{0.098} (70\%) & 67.3
    & \textbf{0.055} (63\%) & 69.4 \\
\bottomrule
\end{tabular}
\end{table}

\begin{table}[t!]
\centering
\caption{\textbf{Wan2.1-14B long-horizon generation.}
Results on 32 held-out MovieGen prompts at 50\,s with the zero-real
closed-loop corrector merged. $\pm$ denotes SEM across prompts.
Dynamic degree checks for motion collapse and is not ranked.}
\label{tab:wan14b_ood}
\footnotesize
\setlength{\tabcolsep}{3.5pt}
\begin{tabular}{@{}lcccccccc@{}}
\toprule
 & \multicolumn{2}{c}{Quality}
 & \multicolumn{2}{c}{Temporal}
 & \multicolumn{3}{c}{Official VBench}
 & Motion \\
\cmidrule(lr){2-3}\cmidrule(lr){4-5}\cmidrule(lr){6-8}
Method & $\Delta$-drift$\downarrow$ & MUSIQ$\uparrow$
       & Lag-2s id.$\uparrow$ & Cuts/video$\downarrow$
       & Subj.$\uparrow$ & Aesth.$\uparrow$ & Imag.$\uparrow$
       & Dynamic \\
\midrule
Adapted base
    & $+15.65{\scriptstyle\pm1.41}$
    & $46.4{\scriptstyle\pm1.5}$
    & 0.791 & 0.375
    & 67.1 & 52.0 & 54.4 & 37.5 \\
\textbf{$+$ \method{}}
    & $\mathbf{+0.97}{\scriptstyle\pm0.50}$
    & $\mathbf{69.0}{\scriptstyle\pm1.3}$
    & \textbf{0.915} & \textbf{0.19}
    & \textbf{86.3} & \textbf{65.7} & \textbf{64.9} & 37.5 \\
\bottomrule
\end{tabular}
\end{table}

\subsection{Independently Trained Base: Causal Forcing}
\label{app:cfbase}
\textbf{Setup.}
To test transfer beyond our causal adaptation, we apply the same recipe to the released chunkwise autoregressive teacher from Causal Forcing~\citep{causalforcing2026}. It shares the Wan2.1-T2V-1.3B architecture and three-latent chunk structure with our base; all 825 checkpoint tensors load strictly. Its independent training uses teacher forcing, so the model sees clean ground-truth history rather than its own generated context. It produces sharp early frames, with first-20\% MUSIQ of about $64$, but its 50\,s rollouts deteriorate within 10--15\,s and reach $\Delta=+11.99$ (Table~\ref{tab:main}). This separates strong short-clip quality from robustness under self-generated history.

\textbf{Protocol.}
Because the release does not specify a complete inference configuration, we evaluate it using our fixed setup: a rolling 21-latent KV window without an attention sink, 20-step UniPC sampling, and seed equal to the prompt index. This differs from its 50-step sampling default. We therefore also test alternative cache configurations on the same 16 prompts (Table~\ref{tab:cfrobust}) to assess how much this choice affects the comparison.

\textbf{Inference robustness.}
A full KV cache without eviction increases drift relative to the rolling window, from $+11.07$ to $+16.34$ on the subset. Attention sinks improve the uncorrected base, reaching $+5.29$ with a half-window sink; context noise worsens it to $+12.59$. Under the rolling-window protocol, the corrector reaches $+3.94$ on those prompts. Combining correction with the half-window sink reduces drift further to $+3.23$. Thus, the correction helps under the fixed protocol and remains useful alongside the strongest cache setting tested. These cache comparisons do not establish the best possible inference configuration for the released checkpoint.

\begin{table}[t!]
\centering
\caption{\textbf{Causal-Forcing inference robustness.}
All configurations use the same 16 prompt–seed pairs at 50\,s;
$\pm$ denotes SEM across prompts. Rolling uses a 21-latent KV window.
Anchoring and cuts are defined in Table~\ref{tab:protocol_metrics}.}
\label{tab:cfrobust}
\small
\begin{tabular}{@{}lcccc@{}}
\toprule
 & \multicolumn{2}{c}{Quality}
 & \multicolumn{2}{c}{Temporal diagnostics} \\
\cmidrule(lr){2-3}\cmidrule(lr){4-5}
Configuration & MUSIQ$\uparrow$ & $\Delta$-drift$\downarrow$
              & Anchoring & Cuts/video$\downarrow$ \\
\midrule
Rolling window (our protocol)
    & $51.5{\scriptstyle\pm2.3}$ & $+11.07{\scriptstyle\pm2.2}$
    & 0.254 & 1.94 \\
Full KV cache (no eviction)
    & $46.4{\scriptstyle\pm3.9}$ & $+16.34{\scriptstyle\pm2.8}$
    & 0.449 & 1.00 \\
Rolling $+$ sink (first chunk)
    & $54.7{\scriptstyle\pm2.6}$ & $+8.48{\scriptstyle\pm1.3}$
    & 0.410 & 2.56 \\
Rolling $+$ sink (half window)
    & $57.2{\scriptstyle\pm3.6}$ & $+5.29{\scriptstyle\pm1.3}$
    & 0.663 & 1.56 \\
Rolling $+$ context noise ($\sigma=0.2$)
    & $47.8{\scriptstyle\pm2.1}$ & $+12.59{\scriptstyle\pm2.2}$
    & 0.180 & 2.75 \\
\midrule
\textbf{$+$ \method{} (our protocol)}
    & $58.8{\scriptstyle\pm3.6}$ & $+3.94{\scriptstyle\pm1.4}$
    & 0.685 & \textbf{0.50} \\
\textbf{$+$ \method{} $+$ sink (half window)}
    & $\mathbf{59.4}{\scriptstyle\pm3.7}$
    & $\mathbf{+3.23}{\scriptstyle\pm0.8}$
    & 0.794 & 0.56 \\
\bottomrule
\end{tabular}
\end{table}

\textbf{Diagnostic and result.}
Before corrector training, the systematic fraction is high across timesteps, $\alpha^*(t)\in[0.93,0.99]$ (Fig.~\ref{fig:alphagate}), and the preliminary drift measurement is $\Delta=+10.83$. Both indicate substantial room for correction. On the full 128-prompt evaluation, one 36-hour run with a one-step stage, one DAgger round, and 600 closed-loop steps reduces drift from $+11.99$ to $+2.80$ ($77\%$) and increases MUSIQ from $52.0$ to $62.1$, without a model-specific hyperparameter search or checkpoint selection.

\begin{figure}[t!]
\centering
\includegraphics[width=0.9\linewidth]{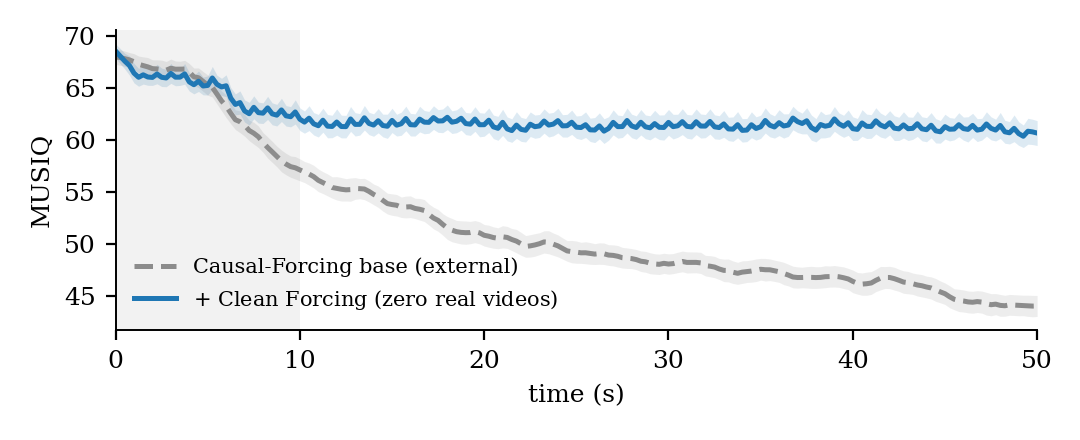}
\caption{\textbf{Quality over 50\,s on the Causal Forcing base.}
Mean per-frame MUSIQ over 128 prompts, smoothed over five frames; bands show SEM. The shaded region marks the first 20\% of each video. Both models start at similar quality, but the base declines throughout the rollout while the corrected model stabilizes after 10\,s.}
\label{fig:cfmoney}
\end{figure}

The corrected checkpoint also improves subject consistency from $64.9$ to $81.8$, background consistency from $75.7$ to $87.9$, aesthetics from $45.8$ to $62.9$, dynamic degree from $33.6$ to $70.3$, and prompt adherence from $57.5$ to $71.0$ (Tables~\ref{tab:main} and~\ref{tab:semantic}). Increased dynamic degree rules out simply freezing the output. The quality curve remains comparatively stable over 50\,s (Fig.~\ref{fig:cfmoney}), while lagged identity improves and the cut rate falls by roughly a factor of three (Table~\ref{tab:protocol_metrics}). Compared with our primary corrected base, this external base retains more residual drift ($+2.80$ versus $+1.65$) and lower absolute MUSIQ ($62.1$ versus $66.9$), so successful transfer does not imply identical final quality.

\textbf{Artifacts across bases.}
The corrected Causal-Forcing rollouts exhibit the same luminance pulse at the three-latent chunk cadence as our primary system (Appendix~\ref{app:limitations}). Periodogram power at that cadence exceeds broadband power in both the uncorrected and corrected runs, with ratios of $309$ and $693$, respectively; neither has a corresponding peak at the 21-latent window cadence. This association with chunk boundaries across two bases suggests an artifact of chunked generation, although correction can make it more visible by reducing background degradation. Anchoring is likewise similar to the primary corrected base (last-fifth DINO similarity $0.680$ versus $0.679$; Table~\ref{tab:protocol_metrics}), while motion energy remains stable from the first fifth to the last in both arms.

On a 16-prompt subset, regenerate-and-blend decoding with two overlap steps adds one-third to decode NFE and lowers the cadence-band power ratio from $538$ to $99$ and autocorrelation from $0.69$ to $0.53$. Motion energy and anchoring remain approximately unchanged. The overlap experiment offers a partial mitigation for the pulse, with additional decode cost.

\begin{figure}[t!]
\centering
\includegraphics[width=\linewidth]{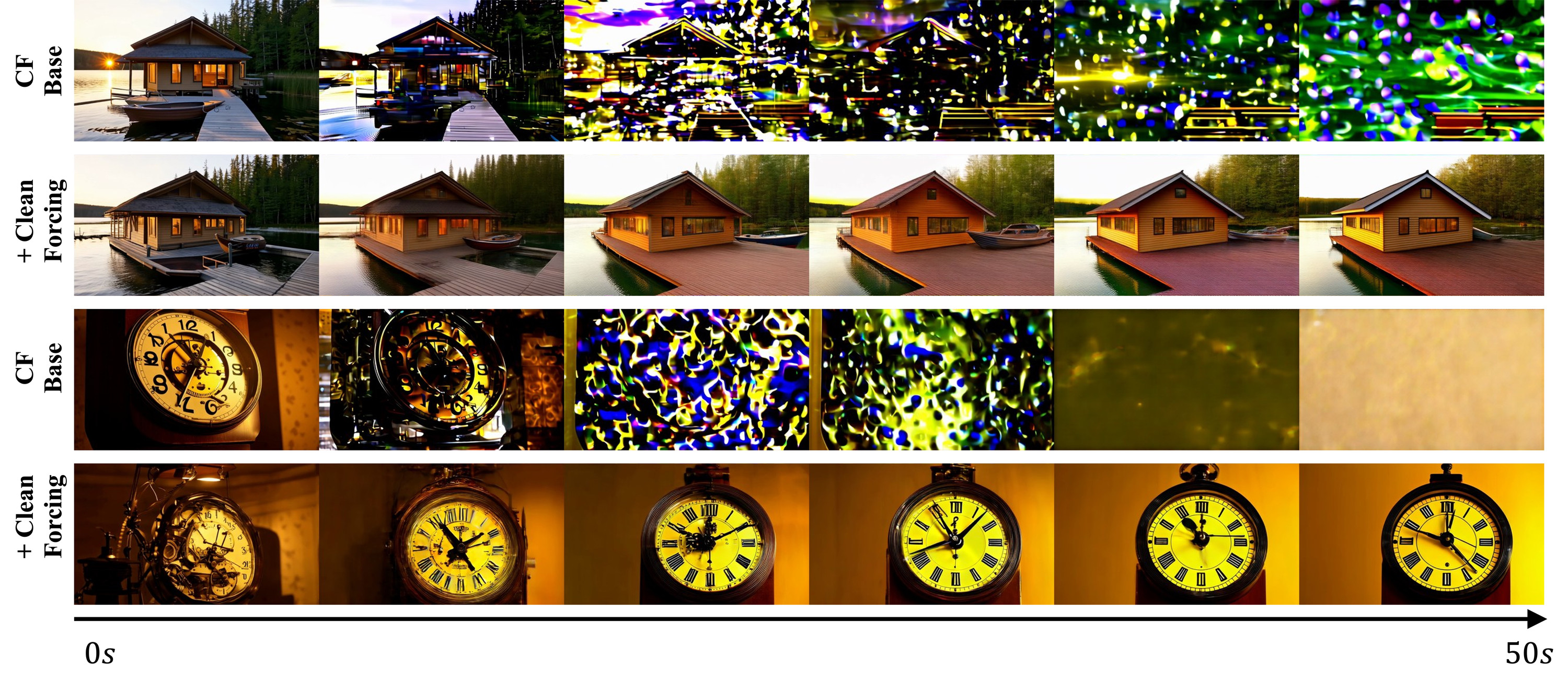}
\caption{\textbf{Qualitative transfer to the Causal Forcing base.}
Six frames sampled over 50\,s for a lakeside cabin (top pair, p040) and a mechanical clock (bottom pair, p101). Each pair uses the same prompt and seed, with the base above and $+$\,\method{} below. The base loses scene structure during rollout, while the corrected model retains recognizable content through 50\,s.}
\label{fig:cfstrip}
\end{figure}

\subsection{Other Autoregressive World Models}
\label{app:otherhosts}
We apply the zero-real-video recipe to two additional autoregressive world models: HY-WorldPlay (5B)~\citep{sun2025worldplay}, which supports interactive camera control, and OmniDreams~\citep{basant2026omnidreams}, a 2B action-conditioned driving model built on Cosmos-Predict2.5~\citep{nvidia2025cosmospredict25}. Unlike our Wan bases, whose systematic fraction $\alpha^*(t)$ remains within $0.89$--$0.99$, both models show substantial variation across timesteps ($0.81\to0.53$ for HY-WorldPlay). We therefore use correction strengths of $0.5\alpha^*(t)$ and $0.25\alpha^*(t)$, respectively. Static-camera HY-WorldPlay rollouts show no drift and are left uncorrected. On held-out rollouts, correction reduces $\Delta$-drift by $48\%$ from $+1.70$ on HY-WorldPlay while preserving dynamic degree ($19.3\to20.3$). On OmniDreams, it reduces $\Delta$-drift by $60\%$ from $+2.44$ and raises late-window MUSIQ by $5\%$, with $17\%$ lower motion. Representative rollouts appear on the project page.

\begin{figure}[t!]
\centering
\includegraphics[width=\linewidth]{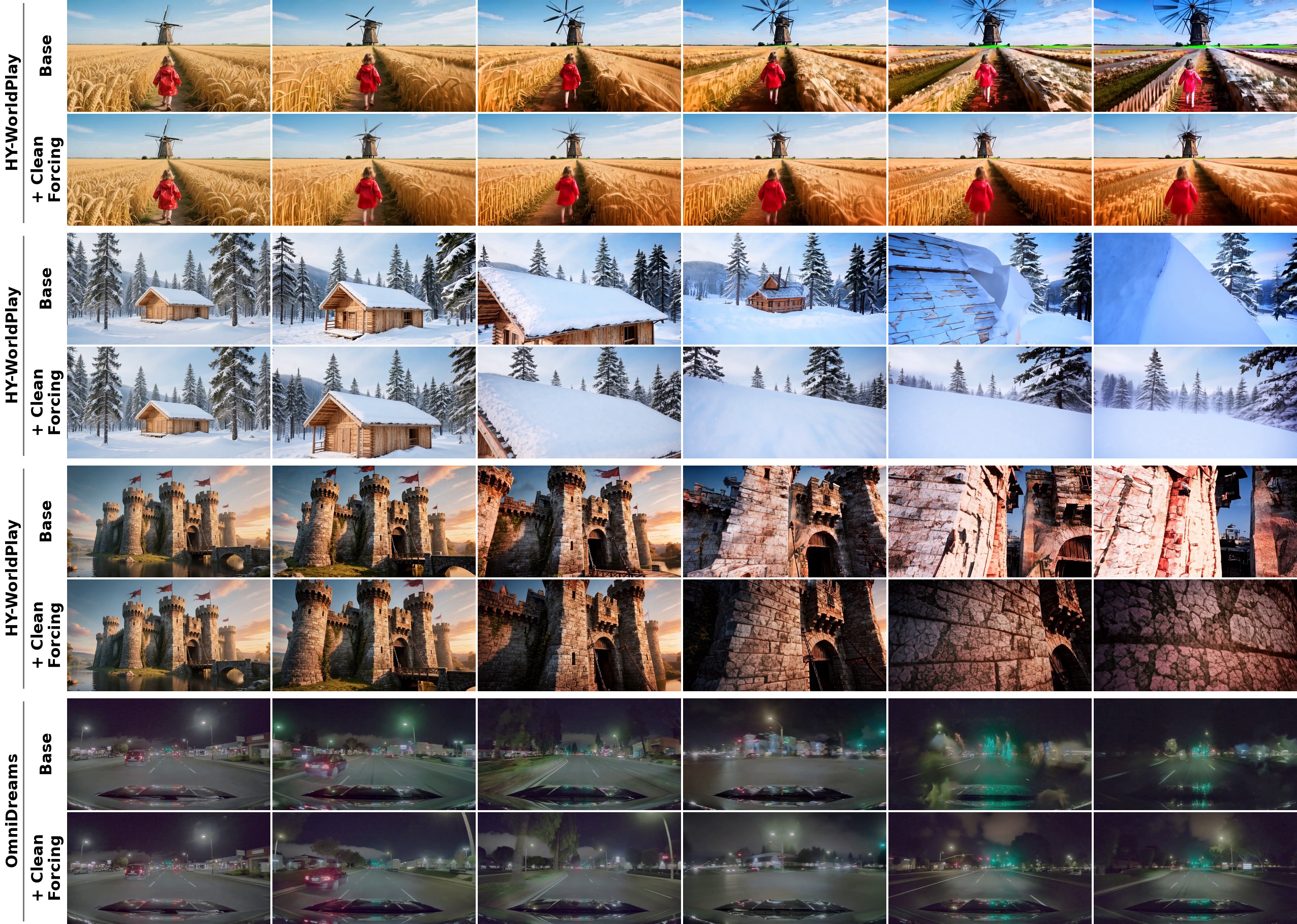}
\caption{\textbf{Correction on other autoregressive world models.} Six evenly spaced frames from three HY-WorldPlay rollouts (${\sim}24$\,s) and one OmniDreams driving rollout (${\sim}80$\,s). Each base and corrected pair shares its prompt, conditioning, and seed. We use strengths of $0.5\alpha^*(t)$ for HY-WorldPlay and $0.25\alpha^*(t)$ for OmniDreams. The corrected rollouts better retain color, detail, and scene structure as the base rollouts degrade.}
\label{fig:otherhosts}
\end{figure}

\section{Extended Ablations}
\label{app:ablations_ext}

\begin{table}[t!]
\centering
\caption{\textbf{Corrector capacity, placement, and rank.}
Paired in-domain evaluation on 48-latent-frame extensions (${\sim}12$\,s).
The upper panel compares corrector parameterizations on the unadapted base;
the lower panel varies LoRA rank on the adapted base. Drift removal is
relative to each panel's base. Bold marks the best scores and the two
MUSIQ losses relative to rank 16; underlining marks second-best scores.}
\label{tab:ablations_ext}
\label{tab:rank}
\small

\begin{tabular}{@{}lcccc@{}}
\toprule
Configuration & Params & Sat. drift$\downarrow$
              & Drift removed$\uparrow$ & Late MUSIQ$\uparrow$ \\
\midrule
Rank-16 LoRA in backbone (\textbf{Ours})
    & 5.9M & \textbf{0.131} & \textbf{73\%} & \textbf{60.5} \\
\quad w/o low rank (full-rank fine-tune)
    & 283M & 0.379 & 23\% & --- \\
\quad w/o in-backbone placement (side network)
    & 1.9M & \underline{0.367} & \underline{25\%} & --- \\
\midrule
Unadapted base & 0 & 0.491 & --- & 29.6 \\
\bottomrule
\end{tabular}

\vspace{8pt}

\begin{tabular}{@{}lccccc@{}}
\toprule
Rank & Params & Val.\ $R^2$
     & Sat. drift$\downarrow$ & Late MUSIQ$\uparrow$
     & $\Delta$MUSIQ vs.\ r16 \\
\midrule
8 & 3.0M & 0.829
   & $0.048{\scriptstyle\pm0.027}$ ($89\%$)
   & $\underline{55.3{\scriptstyle\pm7.2}}$
   & $-1.0{\scriptstyle\pm0.8}$ (n.s.) \\
16 (\textbf{Ours}) & 5.9M & \textbf{0.854}
   & $\underline{0.047{\scriptstyle\pm0.033}}$ ($89\%$)
   & $\mathbf{56.3{\scriptstyle\pm8.5}}$ & --- \\
32 & 11.8M & \underline{0.843}
   & $0.050{\scriptstyle\pm0.012}$ ($88\%$)
   & $49.1{\scriptstyle\pm4.7}$
   & $-7.2{\scriptstyle\pm2.7}$ \\
64 & 23.6M & \underline{0.843}
   & $\mathbf{0.031{\scriptstyle\pm0.015}}$ ($93\%$)
   & $51.7{\scriptstyle\pm6.7}$
   & $-4.7{\scriptstyle\pm1.4}$ \\
\midrule
Adapted base & 0 & ---
   & $0.426{\scriptstyle\pm0.085}$
   & $38.3{\scriptstyle\pm11.7}$ & --- \\
\bottomrule
\end{tabular}
\end{table}

\textbf{Capacity and placement.}
Table~\ref{tab:ablations_ext} changes the corrector architecture while holding the objective, data, and evaluation protocol fixed. Full fine-tuning and a side network each reduce drift by roughly one quarter, compared with $73\%$ for rank-16 LoRA. Full fine-tuning uses 283M trainable parameters and reaches validation $R^2=0.763$, yet gives only a $23\%$ drift reduction in rollout. Its validation fit is lower than the rank-16 LoRA's $0.854$ in the rank sweep (Table~\ref{tab:rank}); the difference in experimental settings should be considered when comparing those values. The 1.9M-parameter side network also underperforms despite having fewer parameters than LoRA, reducing drift by $25\%$. A LoRA modifies the attention projections within the velocity model, whereas the side network adds a residual after those representations are formed. The rank sweep in Table~\ref{tab:rank} further tests how capacity affects the balance between one-step fit and rollout quality.

\textbf{LoRA rank.}
Table~\ref{tab:rank} varies only the rank in a matched one-step experiment on the adapted base. Validation $R^2$ is similar from rank 8 to 64 ($0.829$--$0.854$), indicating that one-step fit largely saturates at low rank. Late-window rollout quality instead peaks at rank 16 in this sweep. Ranks 32 and 64 score $7.2$ and $4.7$ MUSIQ lower than rank 16 on paired clips, without a corresponding improvement in drift. Rank 8 differs from rank 16 by only $1.0\pm0.8$ MUSIQ. These results motivate rank 16 as the deployed setting while showing that more capacity does not necessarily improve long-horizon output.

\section{Additional Results on Drift, Quality, and Diversity}
\label{app:metrics}

\begin{figure}[t!]
\centering
\includegraphics[width=1\linewidth]{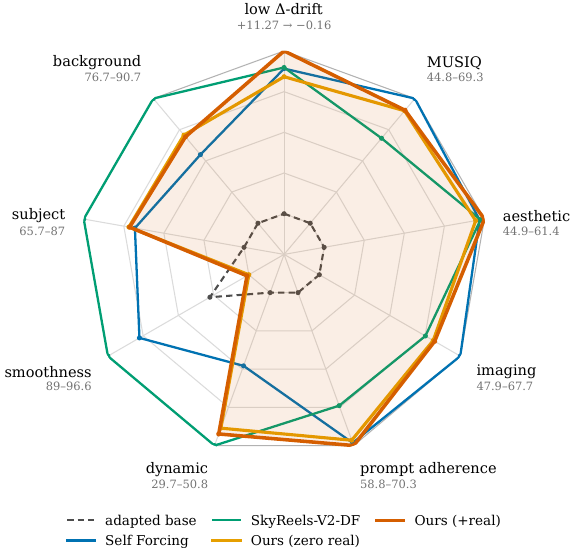}
\caption{\textbf{VBench and drift comparison.}
Scores from Tables~\ref{tab:main} and~\ref{tab:semantic} are normalized by axis; the drift axis is reversed so that outward is better. \method{} improves the adapted base across most measures and is competitive with the retrained models, though it remains lower on smoothness and consistency.}
\label{fig:radar}
\end{figure}

\textbf{Isolating the corrector.}
The paired $2\times2$ experiment separates correction from causal adaptation (Table~\ref{tab:2x2}). Correction reduces saturation drift by $66$--$73\%$ on the unadapted base and $80$--$85\%$ on the adapted base. Adaptation enables prompt-only generation and raises uncorrected late MUSIQ from $29.6$ to $38.5$ in this setting; correction then reduces the remaining drift. On the unadapted base, the closed-loop stage also raises late MUSIQ from $36.9$ after one-step correction to $60.5$. Alternative metrics preserve the principal drift ranking (Table~\ref{tab:suite}). The two corrected variants include the lowest TetherCache $\Delta$QualityDrift ($+0.36$ versus $+4.6$ for Self Forcing), with strong color stability and anchored content survival. Their temporal LPIPS remains relatively poor among outputs that retain motion, consistent with the visible chunk seam.

\textbf{Zero real videos.}
Single-shot bidirectional generations from the base provide the clean references needed to train the causal adapter and corrector using prompts alone (Table~\ref{tab:main}). This variant reaches MUSIQ $66.9$ and $\Delta=+1.65$, compared with $67.0$ and $-0.16$ using 40 real clips. Its drift score is also lower than the published $+3.57^\dagger$ result, which uses a different base and prompt set and is therefore an indicative comparison. Reference-free distribution matching removes no drift in our ablation. The small training set is consistent with a highly systematic velocity gap (\S\ref{sec:measurement}), and drift changes little across the tested 16, 40, and 300 clip regimes.

\textbf{Short-horizon quality.}
The corrector is trained on long rollouts, so we also measure its effect before substantial drift develops. On 200 VBench prompts at 5\,s, the merged corrector improves five of six reported dimensions, including imaging quality ($+4.0$), subject consistency ($+2.3$), and dynamic degree ($+12.5$). Motion smoothness falls by $1.7$ points as motion increases (Table~\ref{tab:dnh}).

\textbf{Inter-seed diversity.}
Correction preserves similar semantic diversity across seeds. DINO inter-seed dissimilarity, measured over three seed pairs, eight extension frames, and five clips, is $0.31$--$0.38$ for the corrected configurations versus $0.33$ for the base, with no consistent trend across ranks or closed-loop training. Pixel-level LPIPS falls by $14$--$18\%$ for the deployed correctors, but seed-dependent saturation artifacts in the base can inflate this pixel-space measure. The DINO comparison therefore provides a complementary check on semantic diversity.

\textbf{Prompt adherence.}
Table~\ref{tab:semantic} adds the VBench temporal-flickering and overall-consistency dimensions available on custom prompts. Overall consistency uses the official ViCLIP video--text similarity against each generation prompt, truncated to 32 tokens. Eight other dimensions require labeled prompts from the official suite, human action relies on a Kinetics label parsed from the filename, and temporal style duplicates overall consistency in this setting. The uncorrected base loses $11.5$ points in overall consistency as it drifts, while the TTC variants score $47$--$50.5$. Correction restores prompt adherence to approximately $70$, close to the distilled models, with the zero-real variant within $0.4$ points of $+$real. Temporal flickering is less informative here: the nearly static history-guidance output scores best among 16\,fps rows, as discussed in Appendix~\ref{sec:guards}.

\textbf{Horizon sweep.}
We truncate the same 128 videos at 10, 30, and 50\,s and score each segment (Table~\ref{tab:lengths}); the 50\,s results match Table~\ref{tab:main}. The base already reaches $\Delta=+10.7$ and MUSIQ $56.1$ at 10\,s, then loses another $11.3$ MUSIQ by 50\,s. The $+$real variant remains stable across horizons, while the zero-real variant shows a smaller, gradual decline. The base's $\Delta$ peaks at 30\,s and then decreases despite continued loss of absolute quality. As the collapsed tail occupies more of the video, a first-versus-overall score can saturate, so $\Delta$ should be reported alongside an absolute quality measure.

\begin{table}[t!]
\centering
\caption{\textbf{In-domain extension.}
Five held-out clips are extended by 48 latent frames (${\sim}12$\,s)
under three paired seeds. Values are means $\pm$ standard deviations
across the three seed means. Bold marks the best mean in each metric.}
\label{tab:2x2}
\small
\begin{tabular}{@{}llcc@{}}
\toprule
Base & Correction & Sat. drift$\downarrow$ & Late MUSIQ$\uparrow$ \\
\midrule
Unadapted & None
    & $0.491\pm0.045$ & $29.6\pm1.5$ \\
          & One-step $\rphi$
    & $0.166\pm0.083$ & $36.9\pm2.1$ \\
          & Closed-loop $\rphi$
    & $0.131\pm0.024$ & $60.5\pm1.2$ \\
\midrule
Adapted   & None
    & $0.415\pm0.017$ & $38.5\pm0.8$ \\
          & One-step $\rphi$
    & $\mathbf{0.061}\pm0.013$ & $59.4\pm0.2$ \\
          & Closed-loop $\rphi$
    & $0.082\pm0.006$ & $\mathbf{63.3}\pm0.5$ \\
\bottomrule
\end{tabular}
\end{table}

\begin{table}[t!]
\centering
\caption{\textbf{Drift diagnostics on 50\,s videos.}
Each configuration uses 32 videos from the evaluation set.
$\Delta$QD denotes $\Delta$QualityDrift. Bold marks the best values
among the four drift measures. Temporal LPIPS and flicker can reward
low motion, as illustrated by the nearly static history-guidance
run$^*$ (Appendix~\ref{sec:guards}).}
\label{tab:suite}
\small
\begin{tabular}{@{}lcccccc@{}}
\toprule
 & \multicolumn{4}{c}{Drift measures}
 & \multicolumn{2}{c}{Frame differences} \\
\cmidrule(lr){2-5}\cmidrule(lr){6-7}
Method & $\Delta$QD$\downarrow$
       & Color L1$\downarrow$ & Color corr.$\uparrow$
       & Survival$\uparrow$
       & t-LPIPS$\downarrow$ & Flicker$\downarrow$ \\
\midrule
Adapted base
    & $+22.9$ & 1.43 & 0.16 & 0.44 & 0.23 & 15.9 \\
Context noise
    & $+31.4$ & 1.44 & 0.18 & 0.45 & 0.19 & 11.1 \\
History guidance$^*$
    & $+36.5$ & 1.25 & 0.15 & 0.37 & 0.12 & 8.2 \\
Self Forcing
    & $+4.6$ & 1.33 & 0.06 & 0.64 & 0.15 & 9.5 \\
\midrule
\method{} (zero real)
    & $+3.5$ & 0.95 & 0.48 & \textbf{0.70} & 0.33 & 22.8 \\
\method{} ($+$real)
    & $\mathbf{+0.36}$ & \textbf{0.85} & \textbf{0.57}
    & 0.69 & 0.59 & 32.8 \\
\bottomrule
\end{tabular}
\end{table}

\begin{table}[t!]
\centering
\caption{\textbf{Short-horizon quality.}
Official VBench scores on 200 prompts at 5\,s. Merged \method{}
improves four of five quality dimensions and increases dynamic degree.
Bold marks the higher quality score; dynamic degree is not ranked.}
\label{tab:dnh}
\small
\begin{tabular}{@{}lcccccc@{}}
\toprule
 & \multicolumn{5}{c}{Official VBench $\uparrow$} & Motion \\
\cmidrule(lr){2-6}
Method & Subj. & Backg. & Aesth. & Imag. & Smooth & Dynamic \\
\midrule
Adapted base
    & 91.6 & 93.1 & 57.0 & 64.9 & \textbf{97.2} & 21.5 \\
$+$ \method{} (merged)
    & \textbf{93.9} & \textbf{94.5} & \textbf{59.0}
    & \textbf{68.9} & 95.5 & 34.0 \\
\bottomrule
\end{tabular}
\end{table}

\begin{table}[t!]
\centering
\caption{\textbf{Prompt adherence and flickering.}
Official VBench custom-input scores on 128 matched 50\,s videos,
reported on the calibrated scale with $\pm$ SEM. Prompt adherence
denotes overall consistency. Flickering is a motion-sensitive diagnostic,
and SkyReels' 24\,fps score is not directly comparable with the 16\,fps
rows (Appendix~\ref{sec:guards}). Bold and underlining mark the two
highest mean prompt-adherence scores.}
\label{tab:semantic}
\small
\begin{tabular}{@{}lcc@{}}
\toprule
Method & Prompt adherence$\uparrow$ & Flickering \\
\midrule
Adapted base
    & $58.81\pm1.32$ & $90.25\pm0.72$ \\
\quad $+$ Context noise ($\sigma=0.2$)
    & $61.68\pm1.33$ & $86.38\pm1.20$ \\
\quad $+$ History guidance ($w=1.2$)
    & $62.15\pm1.08$ & $93.94\pm0.18$ \\
\quad $+$ Pathwise TTC ($\{500,250\}$)
    & $47.00\pm1.52$ & $89.94\pm0.64$ \\
\quad $+$ Pathwise TTC ($\{750,500\}$)
    & $50.53\pm1.55$ & $89.86\pm0.68$ \\
\textbf{\quad $+$ \method{} (zero real videos)}
    & $69.85\pm1.23$ & $87.29\pm0.61$ \\
\textbf{\quad $+$ \method{} ($+$40 real clips)}
    & $\underline{70.28}\pm1.11$ & $86.75\pm0.36$ \\
\midrule
Causal-Forcing base
    & $57.46\pm1.28$ & $88.42\pm1.10$ \\
\textbf{\quad $+$ \method{} (zero real videos)}
    & $\mathbf{70.97}\pm1.11$ & $89.59\pm0.72$ \\
\midrule
Self Forcing
    & $70.12\pm1.19$ & $93.37\pm0.68$ \\
SkyReels-V2-DF
    & $67.31\pm1.29$ & $93.97\pm0.56$ \\
\bottomrule
\end{tabular}
\end{table}

\begin{table}[t!]
\centering
\caption{\textbf{Horizon sweep.}
For each method, the same 128 videos are rescored at 10, 30, and
50\,s ($\pm$ SEM). The base's $\Delta$-drift shrinks after 30\,s even
as MUSIQ falls, so drift should be read alongside absolute quality.
Bold marks the best mean at each horizon for each metric.}
\label{tab:lengths}
\small
\begin{tabular}{@{}llccc@{}}
\toprule
Method & Metric & 10\,s & 30\,s & 50\,s \\
\midrule
Adapted base
    & $\Delta$-drift$\downarrow$
    & $+10.74{\scriptstyle\pm0.69}$
    & $+13.17{\scriptstyle\pm0.89}$
    & $+11.27{\scriptstyle\pm0.81}$ \\
    & MUSIQ$\uparrow$
    & $56.1{\scriptstyle\pm1.2}$
    & $47.4{\scriptstyle\pm1.0}$
    & $44.8{\scriptstyle\pm0.95}$ \\
\addlinespace[2pt]
\textbf{\quad $+$ \method{} (zero real)}
    & $\Delta$-drift$\downarrow$
    & $+1.25{\scriptstyle\pm0.29}$
    & $+1.30{\scriptstyle\pm0.39}$
    & $+1.65{\scriptstyle\pm0.35}$ \\
    & MUSIQ$\uparrow$
    & $68.6{\scriptstyle\pm0.8}$
    & $67.5{\scriptstyle\pm0.8}$
    & $66.9{\scriptstyle\pm0.80}$ \\
\addlinespace[2pt]
\textbf{\quad $+$ \method{} ($+$40 real)}
    & $\Delta$-drift$\downarrow$
    & $+0.39{\scriptstyle\pm0.26}$
    & $\mathbf{-0.01}{\scriptstyle\pm0.16}$
    & $\mathbf{-0.16}{\scriptstyle\pm0.12}$ \\
    & MUSIQ$\uparrow$
    & $66.8{\scriptstyle\pm0.6}$
    & $67.0{\scriptstyle\pm0.6}$
    & $67.0{\scriptstyle\pm0.61}$ \\
\midrule
Causal-Forcing base
    & $\Delta$-drift$\downarrow$
    & $+3.60{\scriptstyle\pm0.42}$
    & $+10.94{\scriptstyle\pm0.66}$
    & $+11.99{\scriptstyle\pm0.67}$ \\
    & MUSIQ$\uparrow$
    & $63.9{\scriptstyle\pm0.8}$
    & $55.7{\scriptstyle\pm0.8}$
    & $52.0{\scriptstyle\pm0.74}$ \\
\addlinespace[2pt]
\textbf{\quad $+$ \method{} (zero real)}
    & $\Delta$-drift$\downarrow$
    & $+2.07{\scriptstyle\pm0.31}$
    & $+3.64{\scriptstyle\pm0.47}$
    & $+2.80{\scriptstyle\pm0.42}$ \\
    & MUSIQ$\uparrow$
    & $64.9{\scriptstyle\pm0.9}$
    & $62.7{\scriptstyle\pm1.0}$
    & $62.1{\scriptstyle\pm0.99}$ \\
\midrule
Self Forcing
    & $\Delta$-drift$\downarrow$
    & $\mathbf{-0.31}{\scriptstyle\pm0.25}$
    & $+0.29{\scriptstyle\pm0.31}$
    & $+1.08{\scriptstyle\pm0.39}$ \\
    & MUSIQ$\uparrow$
    & $\mathbf{70.4}{\scriptstyle\pm0.7}$
    & $\mathbf{70.1}{\scriptstyle\pm0.7}$
    & $\mathbf{69.3}{\scriptstyle\pm0.60}$ \\
\addlinespace[2pt]
SkyReels-V2-DF
    & $\Delta$-drift$\downarrow$
    & $+0.35{\scriptstyle\pm0.22}$
    & $+0.82{\scriptstyle\pm0.27}$
    & $+1.00{\scriptstyle\pm0.30}$ \\
    & MUSIQ$\uparrow$
    & $62.5{\scriptstyle\pm1.0}$
    & $61.9{\scriptstyle\pm1.0}$
    & $61.5{\scriptstyle\pm1.04}$ \\
\bottomrule
\end{tabular}
\end{table}

Table~\ref{tab:protocol_metrics} reports anchoring, two-second lagged identity, cut rate, and intent-conditioned progression excess for all 128 prompts at 50\,s. Scoring every frame rather than using the standard stride changes MUSIQ by at most $1.8$ and $\Delta$ by at most $0.26$, without changing the $\Delta$-drift ranking; only the two \method{} variants swap MUSIQ order, by at most $0.5$. The resulting MUSIQ/$\Delta$ pairs are $65.7/+1.60$ for zero real videos, $65.2/-0.10$ for $+$real, $68.5/+1.34$ for Self Forcing, and $61.5/+0.98$ for SkyReels. We released per-dimension VBench results and per-prompt MUSIQ curves for each configuration in Table~\ref{tab:main}, together with the color-shift, survival, and temporal-LPIPS scores of Table~\ref{tab:suite}.

\begin{table}[t!]
\centering
\caption{\textbf{Scene progression and temporal continuity.}
Results on 128 matched 50\,s videos scored through one pipeline.
Anchoring is DINO similarity between the opening and final fifth;
paired excess subtracts Self Forcing's anchoring prompt by prompt.
Lagged identity uses a 2\,s offset, and cuts are detected between
adjacent frames. Anchoring reflects the drift–progression trade
(Appendix~\ref{sec:trade}).}
\label{tab:protocol_metrics}
\small
\begin{tabular}{@{}lcccc@{}}
\toprule
 & \multicolumn{2}{c}{Opening similarity}
 & \multicolumn{2}{c}{Temporal continuity} \\
\cmidrule(lr){2-3}\cmidrule(lr){4-5}
Method & Anchoring & Excess vs.\ SF
       & Lag-2s identity$\uparrow$ & Cuts/video$\downarrow$ \\
\midrule
Adapted base
    & 0.336 & $-0.148$ & 0.812 & 2.82 \\
\textbf{\quad $+$ \method{} (zero real)}
    & 0.679 & $+0.194$ & 0.830 & 0.89 \\
\textbf{\quad $+$ \method{} ($+$40 real)}
    & 0.795 & $+0.311$ & 0.746 & 0.04 \\
\midrule
Causal-Forcing base
    & 0.249 & $-0.235$ & 0.807 & 1.71 \\
\textbf{\quad $+$ \method{} (zero real)}
    & 0.680 & $+0.196$ & 0.850 & 0.55 \\
\midrule
Self Forcing
    & 0.485 & $0.000$ & 0.922 & 0.26 \\
SkyReels-V2-DF
    & 0.697 & $+0.213$ & 0.927 & 0.12 \\
\bottomrule
\end{tabular}
\end{table}

\section{User Study Details}
\label{app:userstudy}
\textbf{Protocol.}
We use the two-alternative forced-choice format of \citet{yin2025causvid}: raters view two videos side by side with the generation prompt and select the better one. The survey runs on Qualtrics. We choose 26 finals prompts at fixed indices $0,5,\ldots,125$ and compare the $+$real variant with the adapted base, Self Forcing, and SkyReels-V2. Each comparison is rendered with A/B labels, and left--right placement is randomized using a fixed seed. Because our videos last 50\,s, raters view excerpts from $t=40$--$50$\,s, where late drift is visible, rather than the full approximately 10\,s videos used by \citet{yin2025causvid}.

\begin{figure}[t!]
\centering
\includegraphics[width=0.72\linewidth]{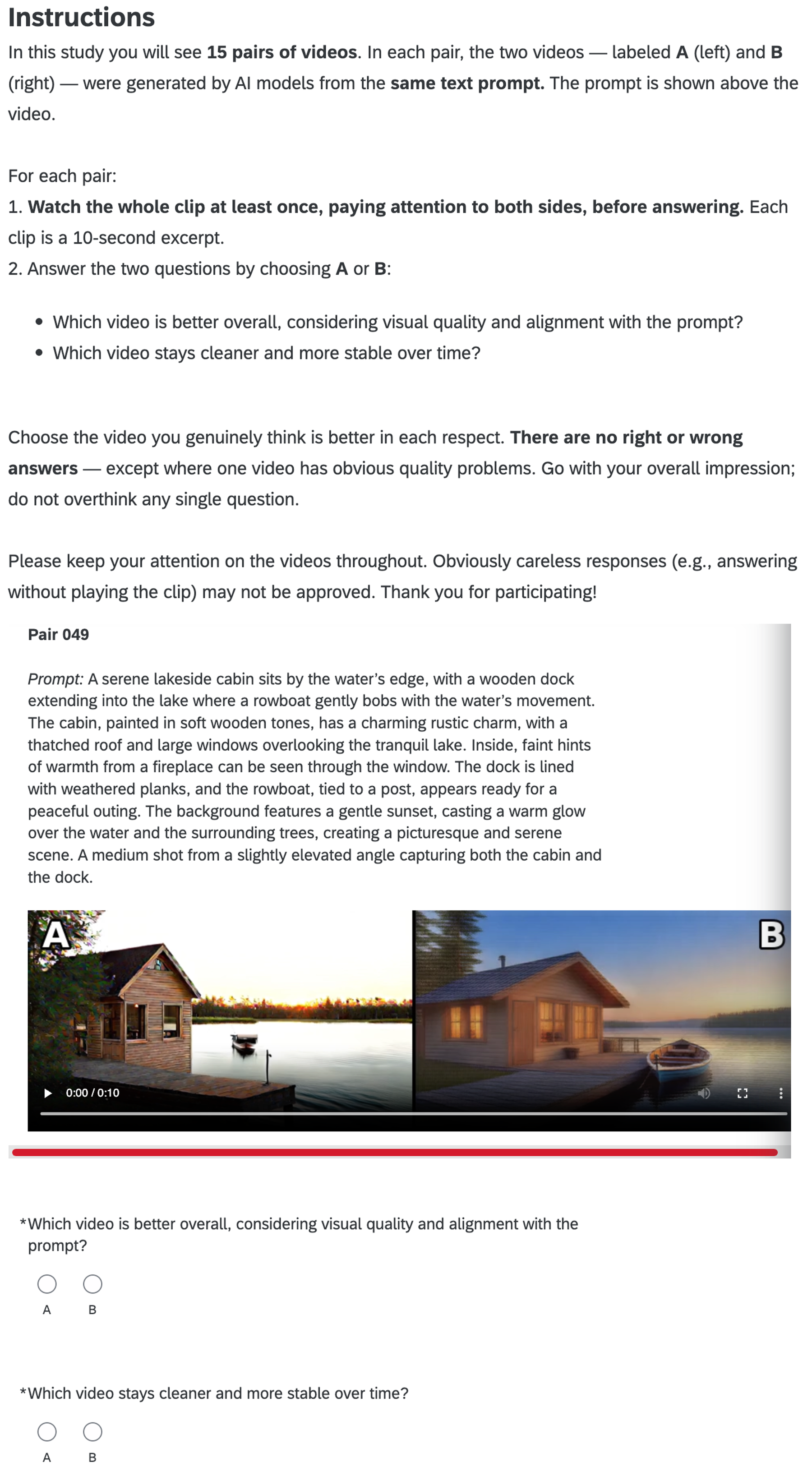}
\caption{\textbf{User study interface.}
Instructions (top) ask raters to watch each 10\,s pair before judging overall quality and temporal stability. A sample trial (bottom) shows the prompt above the two videos, labeled A and B, followed by the two questions.}
\label{fig:userstudy_interface}
\end{figure}

Each rater evaluates a block of 15 pairs: 13 comparisons drawn from the three pairings and two catch trials that compare a healthy video with a collapsed one. The block takes approximately 10 minutes. For each pair, raters answer which video is better overall in visual quality and prompt alignment and which remains cleaner and more stable over time. Figure~\ref{fig:userstudy_interface} shows the instruction page and one trial as raters saw them.

\textbf{Raters and exclusions.}
The analysis includes 33 raters, 29 recruited through Prolific and four directly invited to complete an underfilled block, with at least five ratings per pair. Under the specified catch-failure rule, we excluded one submission that failed both graded catches in its block; we also removed one prelaunch test session. Item analysis identified two of six catch items as nondiscriminative: roughly half of otherwise successful raters failed them, possibly because the collapsed output appeared stylistic. The four retained catch items had failure rates of $0$--$10\%$. Table~\ref{tab:userstudy} reports the resulting preference rates, Wilson intervals, and exact binomial tests.

\begin{table}[t!]
\centering
\caption{\textbf{User study.}
Preference for \method{} ($+$40 real clips) on 10\,s excerpts from
$t=40$--$50$\,s, with 143 judgments per opponent and question.
Brackets give Wilson 95\% intervals; $p$ is the exact two-sided
binomial test against 50\%. Bold marks significant preference for
\method{}.}
\label{tab:userstudy}
\small
\begin{tabular}{@{}llcc@{}}
\toprule
Opponent & Question & \method{} preferred [95\% CI] & Binomial $p$ \\
\midrule
Adapted base & Overall
    & $\mathbf{93.0\%}\ [87.6,96.2]$
    & $1{\times}10^{-28}$ \\
             & Temporal stability
    & $\mathbf{81.8\%}\ [74.7,87.3]$
    & $6{\times}10^{-15}$ \\
\addlinespace[2pt]
Self Forcing & Overall
    & $\mathbf{83.9\%}\ [77.0,89.0]$
    & $5{\times}10^{-17}$ \\
             & Temporal stability
    & $46.9\%\ [38.9,55.0]$
    & n.s. \\
\addlinespace[2pt]
SkyReels-V2-DF & Overall
    & $42.7\%\ [34.8,50.9]$
    & $0.09$ \\
                & Temporal stability
    & $21.0\%\ [15.1,28.4]$
    & $2{\times}10^{-12}$ \\
\bottomrule
\end{tabular}
\end{table}

\begin{figure}[t!]
\centering
\includegraphics[width=1\linewidth]{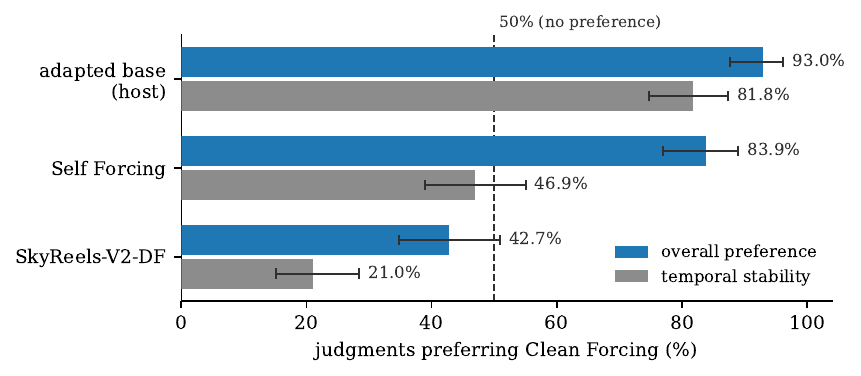}
\caption{\textbf{User study preferences.}
Percentage of judgments favoring \method{} ($+$40 real clips) over each baseline for overall quality (blue) and temporal stability (gray). Whiskers show Wilson 95\% confidence intervals; the dashed line marks 50\% preference (Table~\ref{tab:userstudy}).}
\label{fig:userstudy}
\end{figure}

\textbf{Interpretation.}
The $+$real variant is preferred overall to the adapted base in $93.0\%$ of judgments and to Self Forcing in $83.9\%$. Its $42.7\%$ preference rate against SkyReels-V2 does not differ significantly from an even split ($p=0.09$). On temporal stability, raters prefer our method to the drifting base ($81.8\%$), report no clear preference against Self Forcing ($46.9\%$), and favor SkyReels-V2 ($21.0\%$ for ours). SkyReels-V2 also runs at 24\,fps, compared with 16\,fps for the other evaluated videos, which may affect perceived smoothness. These judgments concern late excerpts and should not be read as whole-video preferences. Rater instructions, anonymized responses, and the analysis script were released with the code.

\section{Additional Qualitative Results}
\label{app:qualitative}
\begin{figure}[t!]
\centering
\includegraphics[width=\linewidth]{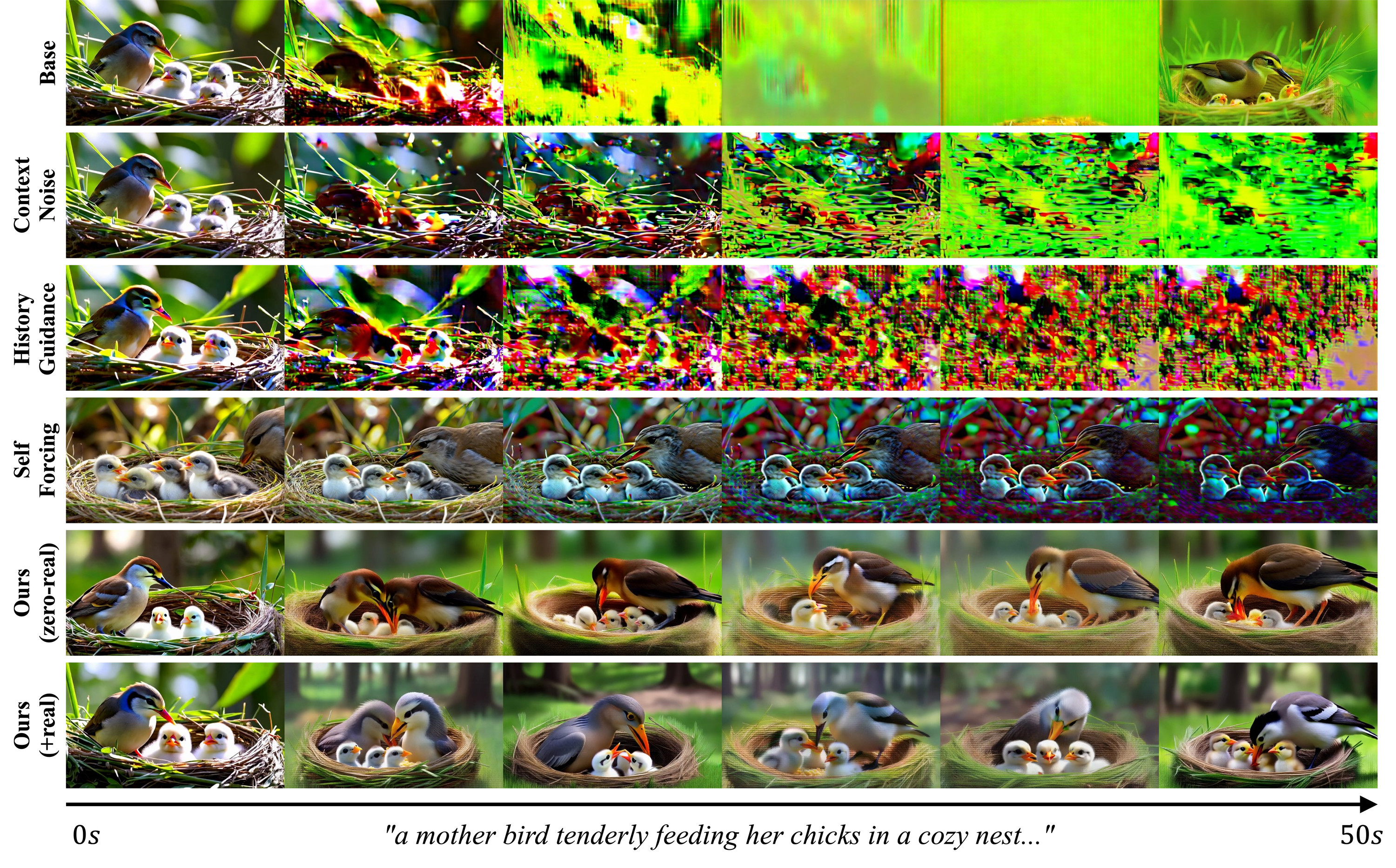}
\caption{Frames from evaluation prompt 113 (mother bird feeding chicks).}
\label{fig:contact2}
\end{figure}

\begin{figure}[t!]
\centering
\includegraphics[width=\linewidth]{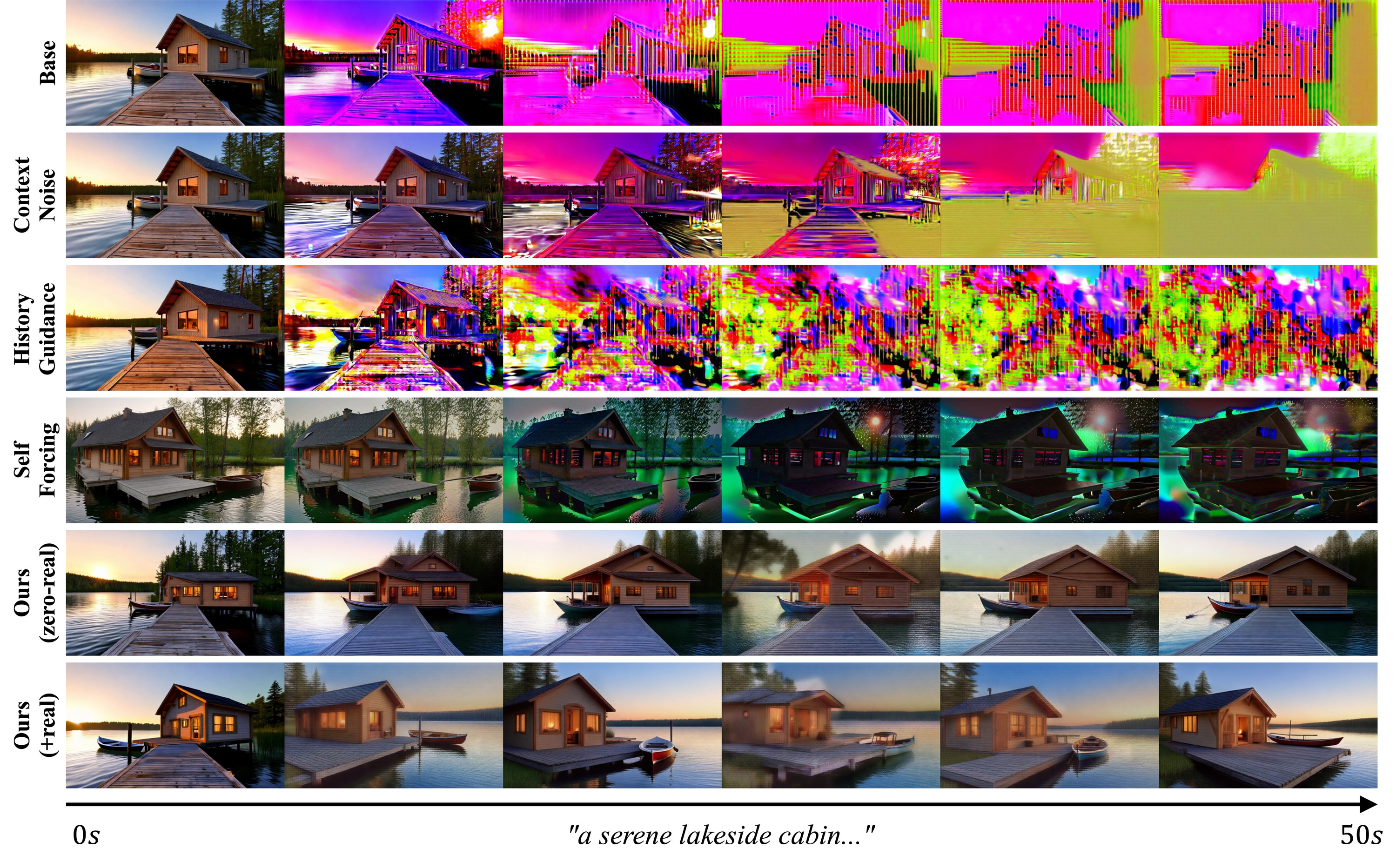}
\caption{Frames from evaluation prompt 40 (lakeside cabin).}
\label{fig:contact3}
\end{figure}

\begin{figure}[t!]
\centering
\includegraphics[width=\linewidth]{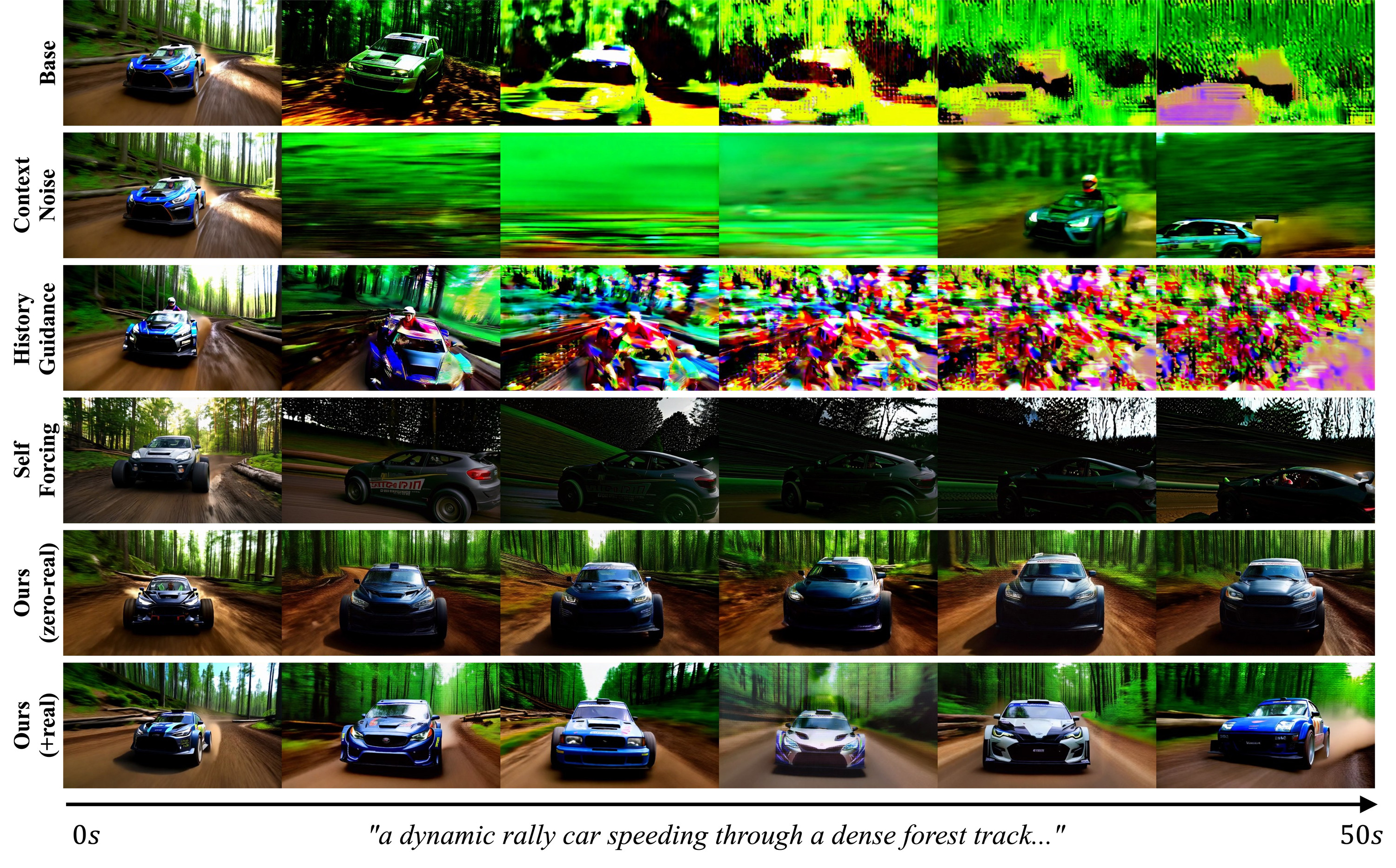}
\caption{Frames from evaluation prompt 8 (rally car through a forest track).}
\label{fig:contact4}
\end{figure}

\begin{figure}[t!]
\centering
\includegraphics[width=\linewidth]{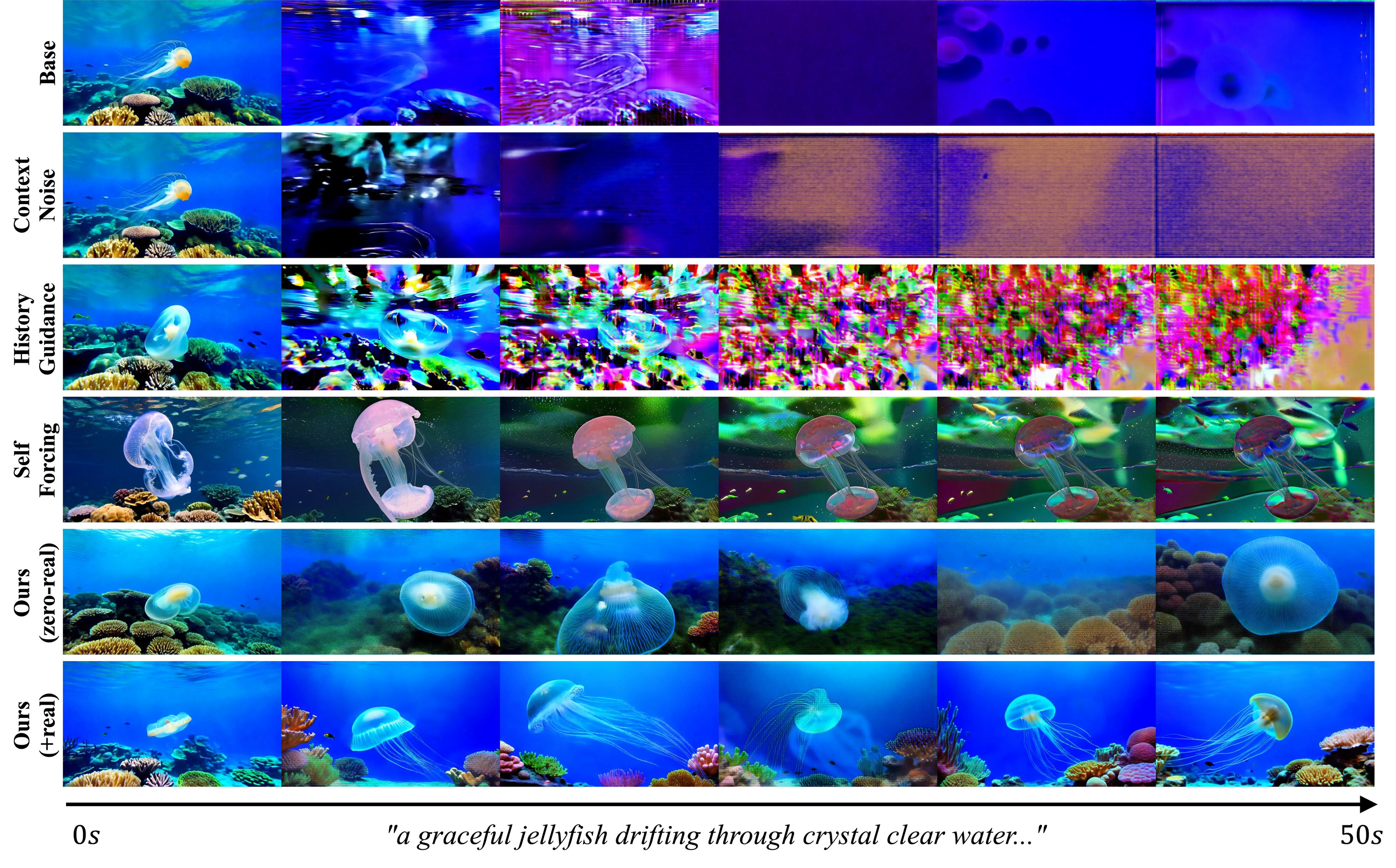}
\caption{Frames from evaluation prompt 28 (jellyfish in clear water).}
\label{fig:contact5}
\end{figure}

\begin{figure}[t!]
\centering
\includegraphics[width=\linewidth]{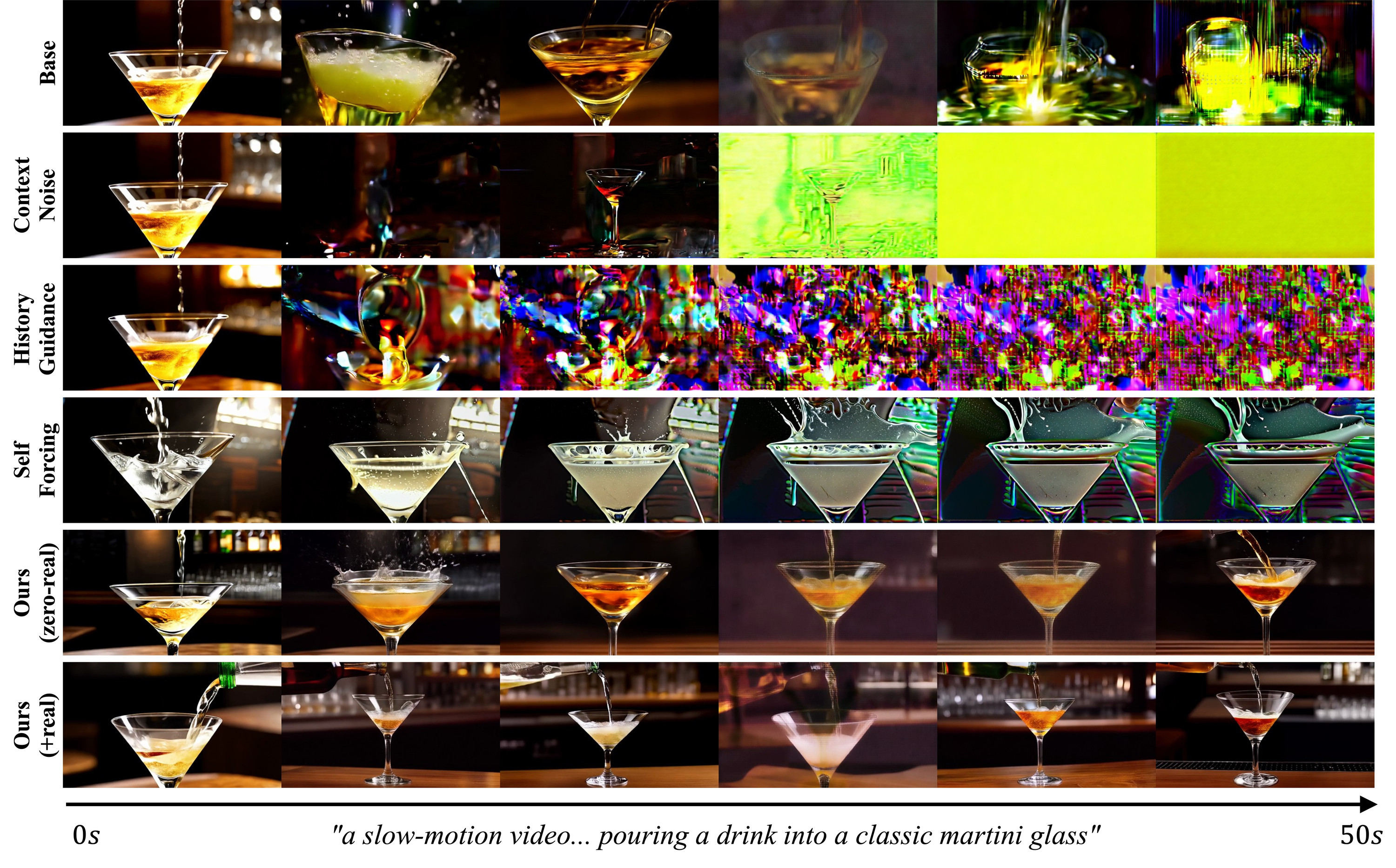}
\caption{Frames from evaluation prompt 32 (martini pour, slow motion).}
\label{fig:contact6}
\end{figure}

\begin{figure}[t!]
\centering
\includegraphics[width=\linewidth]{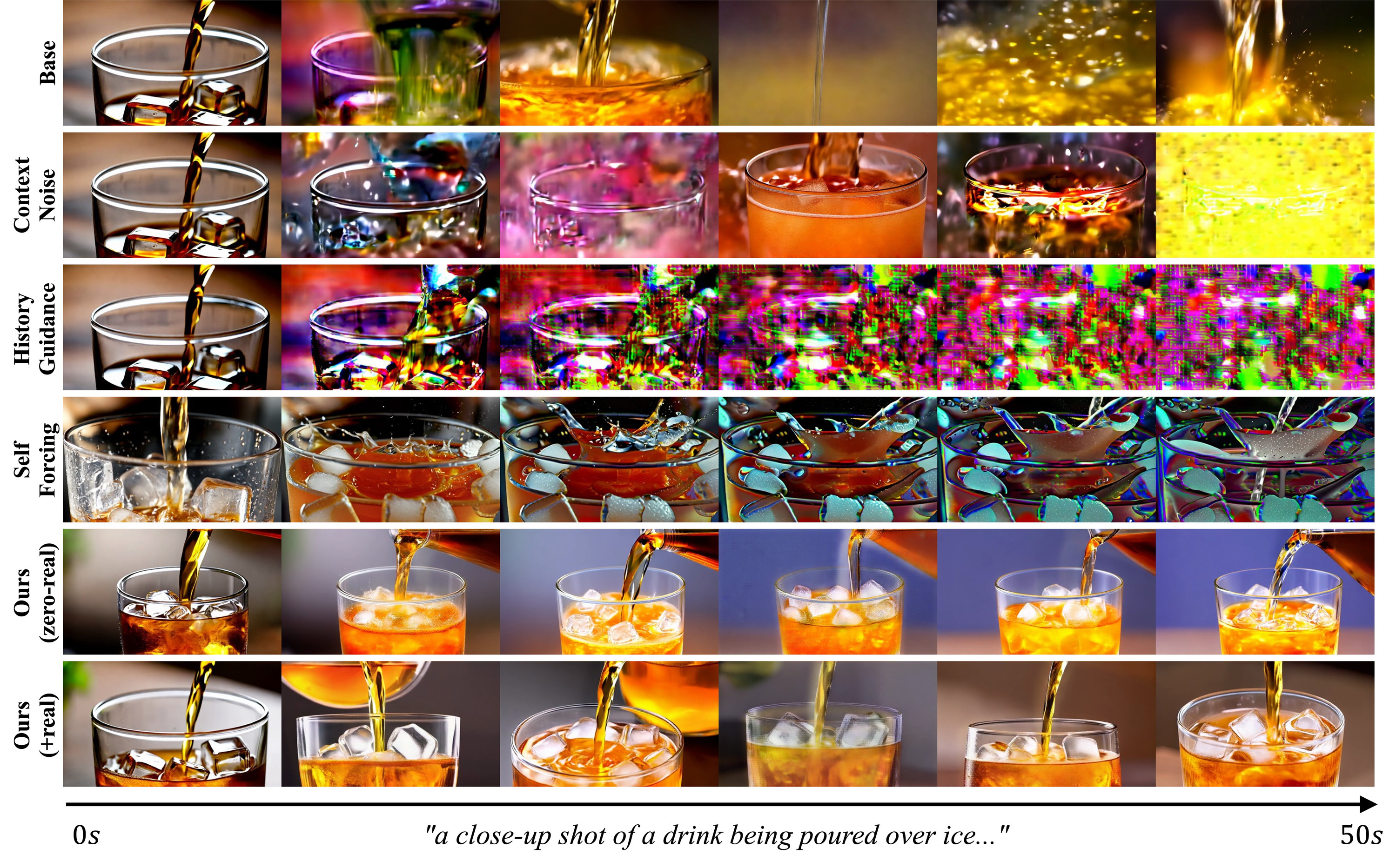}
\caption{Frames from evaluation prompt 30 (drink poured over ice).}
\label{fig:contact7}
\end{figure}

Figure~\ref{fig:qual} in the main text and Figures~\ref{fig:contact2}--\ref{fig:contact7} show evenly spaced frames across 50\,s for eight prompts covering wildlife, vehicles, fluids, and architecture. Seeds are matched across methods for each prompt, and the rows follow the ordering in Figure~\ref{fig:qual}. The project page includes videos for every prompt shown in the figures, including the teaser (Figure~\ref{fig:premise}), method comparisons, and 14B and Causal Forcing examples. It also includes side-by-side clips of the final 10\,s ($t=40$--$50$\,s) from the base and \method{} ($+$real) for all 128 prompts. Full-resolution videos and per-video measurements for all 128 prompts are released with the code.

\clearpage
\section{Limitations}
\label{app:limitations}
Correction substantially reduces quality drift, but three limitations remain. Stronger reference correction can suppress scene progression, a tradeoff the contraction weight adjusts but does not eliminate (\S\ref{sec:trade}). Each new checkpoint or domain requires a new corrector. The base model’s luminance pulse at chunk boundaries also persists (Appendix~\ref{app:cfbase}).

\end{document}